\PassOptionsToPackage{final}{graphicx}
\documentclass[final,5p,times,twocolumn]{elsarticle}

\usepackage{amsmath}            
\usepackage{placeins}

\usepackage[table]{xcolor}   
\usepackage{booktabs}        
\usepackage{pgfplotstable}   
\usepackage{amssymb}
\usepackage{subcaption}
\usepackage{enumitem}

\usepackage{rotating}      
\usepackage{url}
\usepackage{hyperref}

\usepackage{xcolor}

\usepackage{stfloats}      
\usepackage{placeins}      

\usepackage{hyperref}      
\usepackage{svg}  
\usepackage{float}

\usepackage{lineno}

\begin{document}

\begin{frontmatter}

\title{PRISM: Lightweight Multivariate Time-Series Classification through Symmetric Multi-Resolution Convolutional Layers}

\date{March 2026}  

\author[unistra,cephalgo]{Federico Zucchi\corref{cor1}}
\ead{f.zucchi@unistra.fr}

\author[unistra]{Thomas Lampert}
\ead{lampert@unistra.fr}

\cortext[cor1]{Corresponding author}

\affiliation[unistra]{organization={ICube, University of Strasbourg},
              city={Illkirch-Graffenstaden},
              country={France}}
\affiliation[cephalgo]{organization={Cephalgo},
              city={Strasbourg},
              country={France}}

\begin{abstract} 

Multivariate time series classification supports applications from wearable sensing to biomedical monitoring and demands models that can capture multi-scale patterns and dependencies. Despite recent advances, Transformer and CNN models often remain computationally heavy and rely on many parameters. This work presents {PRISM} (Per-channel Resolution Informed Symmetric Module), a lightweight fully convolutional feature extractor. In its early stage it operates in a channel-independent manner, applying a set of multi-resolution symmetric convolutional filters. This symmetry enforces structural constraints inspired by linear-phase FIR filters from classical signal processing, effectively halving the number of learnable parameters within this layer while preserving the full receptive field. Across the diverse UEA multivariate time-series archive as well as specific benchmarks in human activity recognition, sleep staging, and biomedical signals, PRISM matches or outperforms state-of-the-art CNN and Transformer models while using significantly fewer parameters and markedly lower computational cost. By bringing a principled signal processing prior into a modern neural architecture, PRISM offers an effective and computationally economical solution for multivariate time series classification. Code and data are available at 
\href{https://github.com/fedezuc/PRISM}{https:/\allowbreak{}/\allowbreak{}github.\allowbreak{}com/\allowbreak{}fedezuc/\allowbreak{}PRISM}.
\end{abstract}

\end{frontmatter}

\section{Introduction}

Multivariate time series, characterised by intricate temporal dependencies, are common in finance, healthcare, environmental science, and human activity recognition. Deep learning has improved analysis and classification for such data, yet state-of-the-art models often incur high computational cost, heavy parameterisation, and limited robustness in realistic data regimes.

Transformer architectures, adapted from NLP for long-range dependencies, have been applied to time series. Despite promising results, their extensive parameter counts can lead to overfitting and high memory use \cite{wen2023survey}. In practice, self-attention can struggle with noisy, redundant signals \cite{li2022uniformer,wang2025efficient}. Empirical studies also report cases where simple linear or shallow models outperform deep attention architectures on modest-scale datasets \cite{zhang2022less}, suggesting a mismatch between model complexity and many applications.

Conversely, convolutional neural networks (CNNs) have demonstrated considerable success in capturing short- and long-range temporal dynamics via multi-scale architectures \cite{gu2025hdtcnet,mi2022designing}. Even when paired with lightweight classification heads, they can achieve performance comparable to more complex neural architectures, while maintaining a significantly reduced computational footprint \cite{wu2023timesnet}. This motivates a central question: how can one design time-series models that are computationally efficient, compact, and accurate?

In exploring how to design models that remain compact without sacrificing expressiveness, we turned to classical signal processing principles that naturally impose structure and reduce redundant parameterisation. A central question emerged: how can such principles be integrated into modern convolutional architectures for time series. The literature points repeatedly to symmetry as a key structural property. In particular, odd-length symmetric filters behave as linear phase FIR kernels, which preserve waveform shape across frequencies and introduce a strong temporal inductive bias consistent with classical filter design \cite{oppenheim2010discrete}. Symmetry also appears at the heart of many time series tools. Autocorrelation functions are symmetric and capture dependencies that remain invariant under time reversal \cite{papoulispillai2001}. Cosine and several wavelet bases use symmetric patterns to represent oscillatory structure in a compact and phase-consistent way \cite{daubechies1988orthonormal}. Together, these observations show that symmetry supports stable and interpretable representations of temporal dynamics. 

Building on these insights, we introduce PRISM, a compact convolutional module that embeds symmetry and multi-resolution filtering directly into its design. PRISM processes each channel independently using families of symmetric kernels at different temporal scales obtained by mirroring the weights across a central point. This reduces the number of parameters while preserving a rich set of receptive fields. The responses from these filters are then combined through a lightweight mixing stage that produces resolution informed embeddings suitable for classification. This structure enables PRISM to capture multi-scale temporal dynamics. The resulting architecture is simple to train and economical in both memory and computation.

Across diverse benchmarks, PRISM attains competitive or superior accuracy relative to CNN and Transformer baselines while using substantially fewer parameters and compute. Ablations indicate that the per-channel, multi-resolution design and symmetry-based parameter reduction drive favourable accuracy–efficiency trade-offs, and that the approach scales from low- to high-dimensional or multi-modal inputs without specialised fusion blocks.

The contributions of this paper can be summarised as follows:
\begin{itemize}[leftmargin=*, topsep=0pt, itemsep=0pt]

\item \textbf{Parameter efficient multi scale convolution:} We introduce PRISM, a simple and lightweight classifier that at its core employs symmetric kernels constructed by learning half of the filter coefficients and mirroring them around the centre; combined with multiple temporal resolutions, this design reduces the parameter count while maintaining full receptive fields.

\item \textbf{Strong results with high architectural efficiency:} Across UEA, human activity recognition, and biomedical benchmarks, PRISM matches or exceeds the accuracy of considerably larger CNN and Transformer models while operating with markedly fewer parameters and lower computational cost.

\item \textbf{Extensive analysis and flexibility:} Systematic ablations and complexity studies highlight the benefits of the multi resolution structure and show that PRISM scales effectively from low to high dimensional or multi modal inputs under tight memory and FLOP budgets. In addition, empirical analysis on the full UEA collection indicates that symmetric filters learn sharper and more selective frequency responses, stronger stopband attenuation, and greater spectral diversity compared to unconstrained convolutions.

\end{itemize}

\section{Related Works}

\subsection{Symmetric FIR Filters as CNN Feature Extractors}\label{sec:sym_filters}

Recent advances have explored incorporating linear-phase constraints through symmetric (palindromic) finite impulse response (FIR) convolutional filters as initial layers in convolutional neural networks. Such symmetry enforces a linear-phase frequency response, thereby enabling predictable spectral behaviour and facilitating analysis of the frequency characteristics of learned filters \cite{dzhezyan2019symnet}. Specifically, symmetric FIR filters induce a magnitude response equivalent to a linear combination of cosine basis functions, analogous to discrete cosine transforms (DCT) \cite{oppenheim2010discrete,martucci1994symmetric,ahmed1974discrete}, widely appreciated for their spectral compactness in signal compression tasks \cite{ulicny2022harmonic}.

Ulicny et al.\ \cite{ulicny2022harmonic} leverage this spectral relationship by introducing Harmonic Convolutional Networks (HCN), wherein standard convolutional kernels are replaced by learnable weighted combinations of fixed DCT basis functions. This harmonic representation achieves comparable or superior performance to conventional CNNs while substantially reducing redundancy in the learned kernels. Such a structured representation demonstrates improved convergence speeds and enhanced generalisation, attributed directly to the reduced parameter complexity and the robustness of harmonic basis regularisation.

Moreover, within the computer vision domain, SymNet \cite{dzhezyan2019symnet} empirically demonstrates across various image classification benchmarks that imposing strict symmetry constraints on CNN kernels dramatically reduces parameter counts without significant loss of accuracy. Similarly, Liang et al. \cite{liang2021efficient} apply linear-phase constraints to pointwise convolution kernels in efficient image recognition networks, exploiting symmetry across the channel dimension to effectively reduce the parameter count of the projection layers.

The integration of these structured filter constraints serves as a regularisation technique, steering model learning toward frequency-selective, noise-robust feature extraction. Indeed, the empirical advantages reported, mainly in the image domain, in multiple studies--including reduced computational load, accelerated training convergence, improved robustness to overfitting and spectral compactness \cite{ulicny2022harmonic,xu2020learning}--provide strong theoretical and practical justifications for employing such constrained filters as foundational CNN components.

More recently, frequency-constrained and filterbank-style CNN frontends have been proposed specifically for time-series recognition. ASBMNet replaces generic learned kernels with adaptive sinc-like bandpass filters within a multi-scale CNN, targeting interpretable spectral selectivity in underwater acoustic target recognition \cite{qi2026adaptive}. FLEXtime similarly adopts a learned filterbank to decompose signals into frequency bands and to support post-hoc explanations, reinforcing the effectiveness of explicit multi-band frontends for interpretability \cite{brusch2025flextime}.

As such, while prior work incorporates frequency-aware CNNs or filter constraints, to the best of our knowledge, no approach to date explicitly combines per-channel symmetric FIR filtering with multi-resolution design for multivariate time series.

\subsection{Transformer-Based Networks}
Since the seminal Transformer architecture was introduced for NLP \cite{vaswani2017attention}, numerous studies have transplanted self-attention to time–series analysis.  
Early work such as Autoformer \cite{wu2021autoformer}, FEDformer \cite{zhou2022fedformer}, LogSparse/Informer \cite{li2019enhancing}, Reformer \cite{kitaev2020reformer}, and Crossformer \cite{zhang2023crossformer} highlight the ability of attention to model long-range temporal interactions, achieving strong results in multivariate forecasting.  

{Beyond frequency-enhanced attention blocks such as FEDformer, recent work has also explored inserting explicit spectral processing into attention-centric pipelines. Dayag et al.~\cite{dayag2025filter}\ propose a \emph{filter-then-attend} paradigm in which a spectral filtering frontend denoises or reshapes the input prior to a Transformer backbone, suggesting that classical filtering can complement (rather than replace) attention. In a more direct frequency-domain formulation, AttDCT~\cite{haboub2025attdct} performs attention in the DCT domain for time-series classification, further highlighting the utility of cosine-basis representations and spectral symmetry in attention-based models. Moreover, Amplifier \cite{fei2025amplifier} addresses the tendency of existing models to overlook crucial low-energy components in time series forecasting by utilizing a novel energy amplification technique that boosts these neglected signals via spectrum flipping. }

However, recent evidence questions the universal superiority of Transformers for time series.  
Zeng et al.~\cite{zeng2023transformers} demonstrate that the permutation-invariant self-attention mechanism can dilute chronological order, sometimes allowing a one-layer linear model to outperform deep Transformers.    
Moreover, Transformers typically demand large datasets and considerable compute; Wen et al.\ \cite{wen2023survey} report consistent over-fitting on small or noisy benchmarks unless extensive regularisation is used.
These limitations motivate alternatives that embed stronger temporal inductive biases while remaining computationally efficient.

To address these computational overheads and improve representation learning on limited data, recent hybrid architectures have integrated Transformers with convolutional or capsule modules. For instance, DTCM introduces a mutual distillation strategy between a Transformer-based capsule network and a lightweight counterpart to facilitate knowledge transfer \cite{xiao2024dtcm}. Similarly, DKN employs densely dual self-distillation within a hybrid ResNet-Transformer framework to extract a wide variety of constraints \cite{xiao2024densely}, while KATN proposes a knowledge aggregation mechanism to fuse local and global features \cite{xiao2025knowledge}.

\subsection{Convolution-Based Networks}
One-dimensional CNNs excel at learning local motifs in time series and form the backbone of many state-of-the-art classification models.  
ROCKET uses thousands of random convolutional kernels to achieve competitive accuracy with minimal training cost \cite{dempster2020rocket} while subsequent variants like MiniRocket~\cite{dempster2021minirocket} and, more recently, LITE~\cite{ismail2025look} further optimise this efficiency through deterministic features and lightweight architectures.
Self-supervised frameworks such as TS-TCC \cite{eldele2021tstcc}, TS2Vec \cite{yue2022ts2vec}, and MHCCL \cite{meng2023mhccl} rely on CNN encoders to learn transferable representations.

To overcome CNNs' limited receptive field, multi-resolution architectures have proven to be particularly effective \cite{huang2019improved}. SCINet recursively \emph{down-sample–convolve–interact} subsequences, capturing long-term dependencies without attention \cite{liu2022scinet}. Parallel to these structural innovations, recent work incorporates frequency-aware designs. T-WaveNet introduces wavelet-based decomposition to separate low and high-frequency components \cite{minhao2021t}. WFTNet fuses Fourier and wavelet transforms inside a CNN to cover the full temporal spectrum \cite{liu2023wftnet}. TCE regularises 1D-CNNs to focus on informative low-frequency bands \cite{zhang2023temporal}, while BTSF learns bilinear temporal–spectral features for unsupervised representation learning \cite{yang2022unsupervised}. TimesNet reforms forecasting as multi-period analysis in a 2-D space, letting convolutional blocks learn both intra- and inter-period patterns \cite{wu2023timesnet}. Expanding on the modelling of dependencies, SAGOG leverages a similarity-aware graph-of-graphs framework to explicitly capture dynamic correlations between both multivariate channels and time-series instances \cite{wang2025sagog}.

Beyond the specific methods that adopt symmetric filters, as discussed in Section \ref{sec:sym_filters}, recent work has increasingly focused on making the filters themselves adaptive and frequency-aware.
FilterNet employs convolutional kernels that are explicitly trained to act as frequency-selective band-pass filters, enabling the model to match Transformer-level accuracy with far fewer parameters \cite{yi2024filternet}.

These advances show that modern CNNs, augmented with multi-scale processing, spectral priors, and structurally constrained filters, can rival or exceed Transformers in accuracy while remaining more robust on small datasets and dramatically more efficient in FLOPs and memory.

\begin{figure*}[t]
    \centering
    \includegraphics[width=\textwidth]{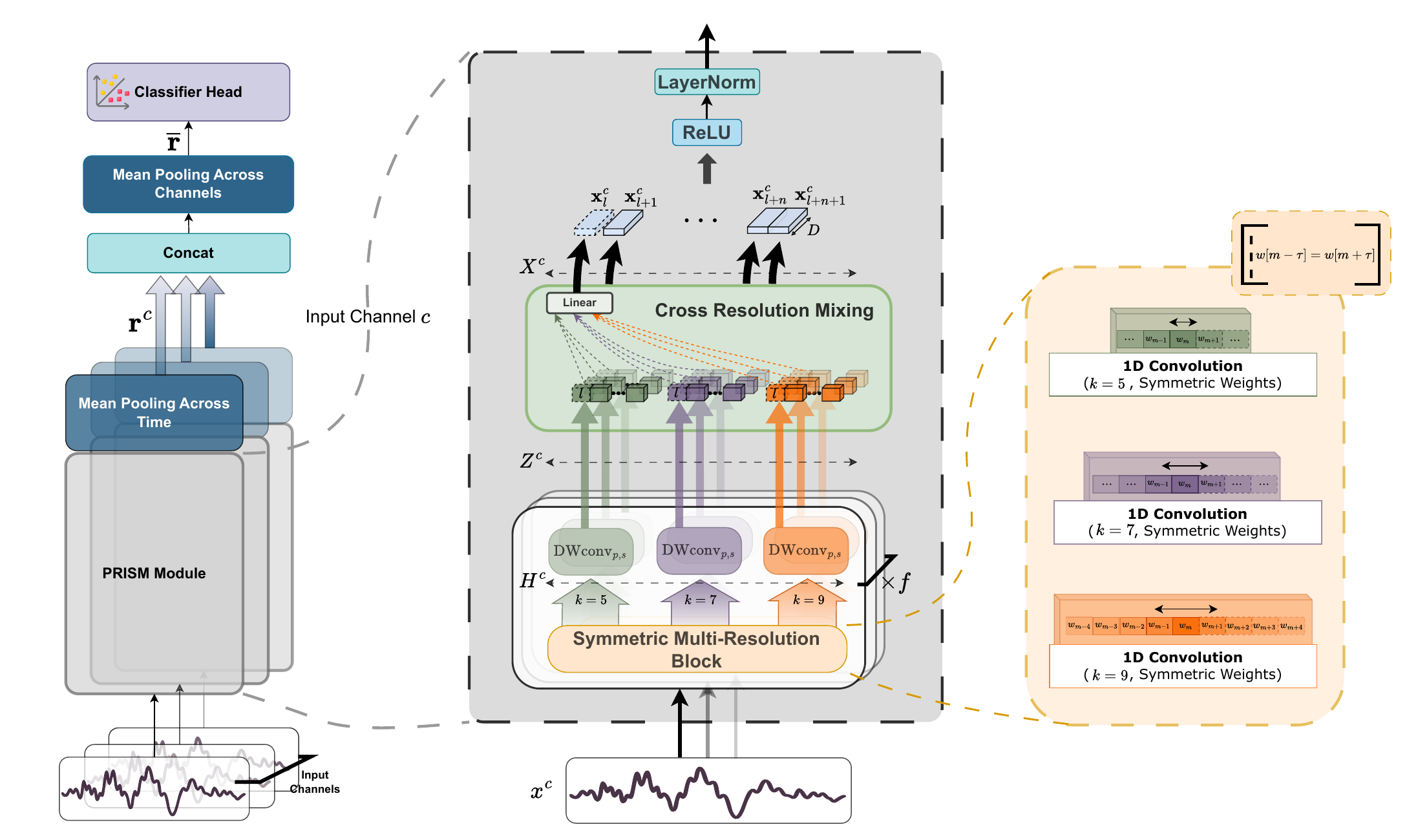} 
    \caption{Starting from the left, the figure shows the full PRISM architecture, where each input channel $\mathbf{x}^{c}$ is processed independently through a shared module. The central panel details this module: a symmetric multi-resolution convolution stage generates resolution specific responses, which are then summarised into local feature sequences $\mathbf{Z}^{c}$. These per-resolution streams are fused by a pointwise mixing layer to form the resolution informed embedding sequence $\mathbf{X}^{c}$ of dimension $D$. This sequence is refined pointwise via ReLU activation to introduce non-linearity, and layer normalisation to standardize the representations. The right panel highlights the symmetric filter design, where kernels are mirrored around their centre $m$ (enforcing $w_{m-\tau}=w_{m+\tau}$) to reduce parameters. After all channels are processed, feature sequences are pooled across time and channels before being passed to a linear classifier.}
        \label{fig:prism-architecture}
\end{figure*}

\section{Methodology}

\subsection{Overview}
\begin{table}[t]
\centering
\caption{PRISM Architectural Terminology}
\label{tab:terminology}
\small
\renewcommand{\arraystretch}{1.3} 
\setlength{\tabcolsep}{3pt}
\resizebox{\columnwidth}{!}{%
\begin{tabular}{@{}lp{0.2\columnwidth}p{0.55\columnwidth}@{}}
\toprule
\textbf{Term} & \textbf{Symbol} & \textbf{Definition and Scope} \\ \midrule
\textbf{PRISM Module} & $-$ & The complete per-channel pipeline from input $x^c$ to representation $r^c$. \\
\textbf{Symmetric Block} & $-$ & Initial module stage using symmetric filters ($w_{m-\tau}=w_{m+\tau}$). \\
\textbf{Resolution Streams} & $H^c$ & Parallel feature maps generated across diverse temporal scales. \\
\textbf{Patch Features} & $Z^c$ & Local feature vectors derived from $H^c$ via depthwise convolution. \\
\textbf{Resolution-Informed Embeddings} & $X^c$ & The fused representation of temporal information after cross-resolution mixing. \\
\bottomrule
\end{tabular}}
\end{table}

PRISM is built around per-channel processing with a symmetric multi-resolution filter module at its core. Table~\ref{tab:terminology} summarises the notation used throughout this section. On the left side of Figure \ref{fig:prism-architecture}, the full architecture is shown: every input channel is transformed independently through the same module, and only then are the resulting representations pooled and combined. This ensures that each channel first develops a resolution-informed embedding before any cross-channel averaging takes place.

The PRISM module, detailed in the central panel of the figure, begins with a Symmetric Multi-Resolution Block. As illustrated in the rightmost breakout panel, the symmetry constraint mirrors each filter around its centre ($w[m-\tau] = w[m+\tau]$), immediately reducing the number of learnable parameters by half while preserving the full receptive field. The use of multiple kernel lengths allows the module to capture structure across a range of temporal extents.

The next stage, labeled $\text{DWConv}_{p,s}$, performs Resolution Specific Feature Extraction. Each resolution stream ($H^c$) is processed independently by a learned depthwise convolutional extractor that summarises short temporal windows into local feature vectors ($Z^c$). This produces local representations that remain faithful to the resolution from which they were derived, with overlapping windows providing controlled downsampling while retaining local context.

These feature sequences are then integrated through Cross Resolution Mixing. A pointwise convolution acts across the resolution dimension at each temporal location, combining the resolution-specific features into a shared embedding space ($X^c$). This operation aligns information from multiple temporal scales within each channel, allowing the model to represent fast and slow dynamics.

Up to this point the transformations are linear, so the resulting feature vectors are passed through a nonlinearity to introduce nonlinear modelling capacity, followed by normalisation to prevent high-energy resolution streams from dominating the fused embedding. After every channel has been processed in this manner, their temporal features are averaged over time, pooled across channels, and passed to a linear classifier. This keeps the final stage simple while ensuring that each channel contributes a well-structured, resolution-aware representation to the final decision. 

Crucially, while the initial symmetric filtering is time-reversal invariant, the subsequent overlapping patch extraction and pointwise nonlinearities (ReLU) break this linearity. This sequence of operations allows the module to retain temporal order and successfully compose asymmetric, directional temporal dynamics (e.g., distinguishing a rising from a falling trend).
Each of these stages is formally defined in the following subsections.

\subsection{Multi-Resolution Symmetric Module}\label{subsec:mksce}
Given an input channel $\mathbf{x}^{c} \in \mathbb{R}^{T}$, we apply a set of 1D convolutional filters with varying odd lengths $\{k_1, \dots, k_{n_k}\}$ and $n_f$ repetitions per size. This results in a total of $C_{\text{res}} = n_k \times n_f$ feature maps, which we refer to as \textit{resolution streams}.
Each kernel $\mathbf{w} \in \mathbb{R}^{k}$ (where $k$ is odd) satisfies the symmetry constraint around its centre index $m = \lfloor k/2 \rfloor$:
\begin{equation}
    w_{m-\tau} = w_{m+\tau}, \quad \forall \tau \in \{1, \dots, m\}.
\end{equation}
This halves the number of learnable parameters while preserving the full receptive field.

Symmetry also induces linear-phase behavior: when the filter is centred at a time step, the contributions of future and past samples are weighted equally. This ensures that all frequency components are delayed uniformly, preventing temporal distortion \cite{oppenheim2010discrete}. 
In particular, the symmetric kernels employed in PRISM correspond to Type I linear-phase FIR filters (real-valued, symmetric impulse response, and odd length). A formal mathematical derivation demonstrating how this symmetry constraint exactly halves the number of independent learnable parameters while maintaining the full receptive field is provided in ~\ref{appendix:sym_filter_theory}.

To maintain numerical stability and preserve frequency characteristics, we apply $\ell_2$-normalisation to the weights before convolution:
\begin{equation}
  \hat{\mathbf{w}} = \frac{\mathbf{w}}{\lVert \mathbf{w} \rVert_2 + \varepsilon}.
\end{equation}
This enforces unit energy, $\lVert \hat{\mathbf{w}} \rVert_2 \approx 1$, decoupling the filter's gain from its shape during gradient updates. The output of this module is a multi-resolution feature tensor $\mathbf{H}^{c} \in \mathbb{R}^{T \times C_{\text{res}}}$, which captures diverse temporal scales within the single input channel $c$.

\subsection{Resolution-Informed Patch Embedding}\label{subsec:patch}
Building upon the multi-resolution features $\mathbf{H}^{c}$, this stage aggregates temporal information into a more compact sequence of local embeddings. We preserve the multi-resolution structure established by the filters while reducing temporal resolution.

\paragraph{Resolution-Specific Patch Extraction}
To capture local temporal structure, we apply a 1D depthwise convolution with kernel size $p$ and stride $s = p/2$, denoted as $\mathrm{DWConv}_{p,s}$. This operation is applied independently to each resolution stream of $\mathbf{H}^{c}$:
\begin{equation}
 \mathbf{Z}^{c} = \mathrm{DWConv}_{p,s}\left( \mathbf{H}^{c} \right),
\end{equation}
where $\mathbf{Z}^{c} \in \mathbb{R}^{L \times C_{\text{res}}}$ is the resulting downsampled feature sequence composed of $L$ steps.
The stride $s$ introduces a controlled overlap between neighboring patches, capturing dependencies that span across adjacent windows. This acts as a learnable pooling operation that refines features within each resolution scale while downsampling.

\paragraph{Cross-Resolution Mixing}
To integrate information across the different resolution streams, we project the features into a shared embedding space of dimension $D$. We apply a learnable linear transformation across the resolution dimension:
\begin{equation}
\mathbf{X}^{c} = \mathbf{Z}^{c} \mathbf{W}_{\text{mix}},
\end{equation}
where $\mathbf{W}_{\text{mix}} \in \mathbb{R}^{C_{\text{res}} \times D}$ is the projection matrix.
Crucially, this operation acts pointwise along the sequence: it linearly combines the $C_{\text{res}}$ resolution streams at sequence index $\ell$ without aggregating information from different steps. The resulting tensor $\mathbf{X}^{c} \in \mathbb{R}^{L \times D}$ forms the resolution-informed embeddings.

\subsection{Nonlinear Projection and Token-Wise Normalisation}
\label{subsec:normrelu}
The embeddings $\mathbf{X}^{c}$ represent the fused representation of the temporal information. Before passing them to downstream components, we apply a sequence of pointwise operations---nonlinearity, dropout, and normalisation---to refine the tokens.

Since all preceding operations are linear transformations, the embeddings $\mathbf{X}^{c}$ currently remain linear combinations of the input time series. To introduce the modelling capacity required for capturing complex non-linear dynamics, we apply a pointwise activation function followed by dropout regularisation:
\begin{equation}
  \mathbf{X}_{\text{act}}^{c} = \mathrm{Dropout}\left( \mathrm{ReLU}\left( \mathbf{X}^{c} \right) \right).
\end{equation}
The ReLU breaks the linearity of the pipeline while dropout reduces co-adaptation among latent dimensions to prevent overfitting. Then, we apply Layer Normalisation (LN) across the embedding dimension $D$:
\begin{equation}
  \mathbf{X}_{\text{out}}^{c} = \mathrm{LN}(\mathbf{X}_{\text{act}}^{c}).
\end{equation}
It stabilises the scale of the final embeddings, ensuring they lie in a standardised latent space before the subsequent pooling.

\subsection{Global Aggregation and Classification}\label{subsec:pooling}
Finally, we aggregate the feature sequences from all input channels into a fixed-dimensional vector for classification.
First, we perform Global Average Pooling over the temporal dimension. For each input channel $c$, let $\mathbf{x}_{t}^{c} \in \mathbb{R}^D$ denote the feature vector at time step $t$ within the sequence $\mathbf{X}_{\text{out}}^{c}$. We compute the channel-specific representation $\mathbf{r}^{c}$ by averaging over the $L$ time steps:
\begin{equation}
\mathbf{r}^{c} = \frac{1}{L} \sum_{t=1}^{L} \mathbf{x}_{t}^{c},
\end{equation}
yielding a vector $\mathbf{r}^{c} \in \mathbb{R}^{D}$. We then average these representations across the $C$ input channels to obtain a global multivariate embedding $\bar{\mathbf{r}}$:
\begin{equation}
  \bar{\mathbf{r}} = \frac{1}{C} \sum_{c=1}^{C} \mathbf{r}^{c}.
\end{equation}
A linear classification head then maps $\bar{\mathbf{r}}$ to class logits via an affine transformation. This hierarchical pooling strategy effectively aggregates local temporal details into a global context before fusing information across multivariate dimensions.

\subsection{Channel Independent Hypothesis}
From a computational perspective, our Channel Independent (CI) strategy is motivated by the critical need for linear scalability and computational efficiency. Channel Dependent (CD) architectures often incur quadratic complexity regarding the channel count, which becomes computationally prohibitive on high-dimensional datasets~\cite{ekambaram2023tsmixer}. By treating channels as independent univariate streams, the CI framework decouples model complexity from the number of variates, preventing the memory bottlenecks associated with dense cross-channel attention mechanisms~\cite{yuqietal2023PatchTST}.
This efficiency is not merely a computational convenience but a structural regularisation. Recent empirical findings from the TSelect~\cite{nuyts2025tselect} algorithm reveal that a significant proportion of channels in multivariate benchmarks are redundant or irrelevant. Complementing this, evaluations of the ECS and ECP algorithms demonstrated that classifier-agnostic channel selection can reduce data storage and computation time by approximately 70\% (specifically reporting around 75\% reduction in runtime and  60\% in memory) while maintaining or even improving accuracy~\cite{dhariyal2023scalable}. These results imply that modelling exhaustive cross-channel interactions in time seriers classification often introduces interaction noise rather than useful signal. This aligns with theoretical analyses by Han et al.~\cite{han2024capacity}, who argue that Channel Dependent (CD) models risk overfitting to spurious correlations in high-dimensional data, whereas CI strategies act as a regulariser by ignoring these unstable dependencies. In fact, state-of-the-art architectures like MD-Former explicitly incorporate a CI branch to prevent the performance degradation caused by ``excessive fusion'' in complex multivariate tasks~\cite{du2025md}.
While we recognise a recent evolving body of work that attempts to reconcile the trade-offs between CI and CD through hybrid architectures \cite{qiu2025comprehensive}, we see in CI an efficient and well-motivated choice for PRISM. This decision is grounded in its structural robustness against distribution drift and linear scalability.

\subsection{Model Complexity Analysis}\label{subsec:complexity}
We characterise the computational and memory complexity of PRISM by decomposing its operations into distinct stages. Let $B$ denote the batch size, $C$ the number of input channels, $T$ the sequence length, $n_k$ the number of odd kernel sizes $\{k_1,\dots,k_{n_k}\}$, and $n_f$ the number of symmetric filters per size. The total number of filters is $C_{\text{res}} = n_k n_f$, with average kernel length $\bar{k} = \frac{1}{n_k}\sum_{s=1}^{n_k} k_s$. For the patch embedding, the number of overlapping patches is $L \approx \frac{2T}{p}$, where $p$ is the patch size (stride $p/2$), and $D$ is the embedding dimension.

\begin{description}[leftmargin=*,topsep=0pt,itemsep=0pt]
  \item[\emph{Filter--Set Stage.}]
  Each input channel is convolved with $C_{\text{res}}$ depth-wise symmetric filters of length $\bar{k}$, requiring $O(B\,C\,C_{\text{res}}\,T\,\bar{k})$ multiply–accumulate operations and $O(C\,C_{\text{res}}\,\bar{k})$ parameters.
  
  \item[\emph{Patch Embedding Stage.}]
  Each filtered channel is divided into $L$ patches. For each patch, a depth-wise convolution (length $p$) followed by a pointwise ($1{\times}1$) projection to $D$ dimensions results in a computational cost of $O(B\,C\,C_{\text{res}}\,L\,(p + D))$ and a parameter count of $O(C_{\text{res}}(p+D) + D)$ per channel.
  
  \item[\emph{Pooling, Normalisation, and Output.}]
  Pooling over $L$ patches incurs $O(B\,C\,D\,L)$ operations, while LayerNorm adds $O(C\,D)$ parameters. The output dimensionality is $(B,C,D)$.
\end{description}

\noindent
All principal hyperparameters ($C$, $C_{\text{res}}$, $\bar{k}$, $T$, $p$, $D$) influence computational and parameter complexity linearly. The scaling of $p$ and $L$ is inversely proportional, while $C_{\text{res}}$ and $\bar{k}$ directly affect temporal scale and cost. The architecture therefore enables precise, linear control over resource allocation and predictable trade-offs between efficiency and representational capacity.

\section{Experiments on Benchmark Datasets}

\subsection{Datasets}\label{subsec:datasets}
To evaluate our model, we adopted datasets from the University of East Anglia (UEA) Time Series Classification repository \cite{bagnall2018uea}, alongside a suite of publicly available datasets from two representative application domains: biomedical signals and human activity recognition (HAR).

While the UEA archive has become a standard benchmark for multivariate time--series classification, several studies have noted limitations when using it to evaluate deep learning models. Prior work has highlighted that many UEA datasets are relatively small and unbalanced, which can restrict the generalisation ability of neural networks. Furthermore, differences in accuracy between deep architectures and more classical approaches (e.g.\ DTW) on these benchmarks are often narrow \cite{ismail2020inceptiontime,forestier2019deep}.

To address these limitations and provide a more realistic assessment of model performance in practical settings, we additionally evaluate PRISM on a selection of biomedical and HAR datasets. These datasets offer larger sample sizes and more controlled class balance compared to the standard UEA benchmarks. Collectively, the selected datasets span a wide range of signal lengths, channel counts, and training sample sizes, providing a comprehensive testbed for evaluating model capabilities. A summary of the main characteristics of the biomedical and HAR datasets is reported in Table \ref{tab:datasets_statistics}. 

\begin{description}[leftmargin=*,topsep=0pt,itemsep=0pt]

\item[\textbf{UEA Time Series Classification Archive}] \cite{bagnall2018uea} provides a large and diverse benchmark collection for time series classification. We tested 29 multivariate datasets covering heterogeneous domains such as human activity recognition, sensor-based systems, environmental monitoring, and biomedical signals. This variety allows for assessing the model’s generalisation ability across different temporal dynamics and data modalities.

\item[\textbf{Sleep-EDF}] sourced from PhysioBank \cite{goldberger2000physiobank}, contains whole-night polysomnography (PSG) recordings used for sleep stage classification. Specifically, EEG signals from the Fpz-Cz channel were extracted at a sampling rate of 100 Hz, consistent with previous studies \cite{eldele2021attention,eldele2021tstcc,eldele2024tslanet}. The classification task includes five sleep stages: Wake (W), Non-Rapid Eye Movement (N1, N2, N3), and Rapid Eye Movement (REM).

\item[\textbf{MIT-BIH Arrhythmia (ECG)}] \cite{moody2001impact}, hosted on PhysioNet, consists of electrocardiogram (ECG) recordings curated for arrhythmia detection and classification. It is a standard benchmark in cardiovascular research, offering a rigorously annotated collection of heartbeats and arrhythmic events widely used for algorithm validation.

\item[\textbf{UCI HAR}] \cite{anguita2013public} contains sensor readings from 30 subjects performing six daily activities (walking, walking upstairs, walking downstairs, standing, sitting, and lying down). Data was recorded with a smartphone (Samsung Galaxy S2) worn at the waist, using accelerometer and gyroscope sensors sampled at 50 Hz.

\item[\textbf{WISDM}] \cite{kwapisz2011activity} includes accelerometer and gyroscope signals from smartphones and wearable devices during activities such as walking, jogging, sitting, and standing. It is commonly used to assess robustness across motion patterns and sensor placements.

\item[\textbf{HHAR}] \cite{stisen2015smart} provides data collected from multiple devices (smartphones and smartwatches) while subjects performed activities such as biking, sitting, standing, walking, and stair climbing. Its heterogeneous device configurations make it a challenging benchmark for evaluating cross-sensor generalisation.
\end{description}

\begin{table}[tb]
\centering
\caption{Dataset statistics: number of training and testing samples, sequence length, number of channels, and classes.}
\label{tab:datasets_statistics}
\resizebox{1\linewidth}{!}{
\begin{tabular}{lccccc}
\toprule
\textbf{Dataset} & \textbf{\# Train} & \textbf{\# Test} & \textbf{Length} & \textbf{\# Channels} & \textbf{\# Classes} \\
\midrule
UCIHAR    & 7,352  & 2,947  & 128   & 9  & 6 \\
WISDM     & 4,731  & 2,561  & 128   & 3  & 6 \\
HHAR      & 10,336 & 4,436  & 128   & 3  & 6 \\
Sleep EDF & 25,612 & 8,910  & 3,000 & 1  & 5 \\
MIT-BIH Arrhythmia (ECG)       & 87554 & 21892   & 187   & 1  & 2 \\
\bottomrule
\end{tabular}
}

\end{table}

\subsection{Experimental Setup}

To ensure a fair and consistent comparison, all models were trained within the same experimental framework using an initial learning rate of $1\times10^{-3}$ and a step-decay schedule that halves the learning rate after each epoch. Training was conducted with the RAdam optimizer, which provides an adaptive learning rate mechanism with variance rectification to improve stability during the early optimisation phase, and the cross-entropy loss function was used as the objective criterion for all classification tasks. To maintain comparable representational capacity across architectures, all models were configured with a fixed embedding size of 128. Model selection was based on the highest validation accuracy, and to account for randomness in initialisation, each experiment was repeated three times with different random seeds; we report the mean test accuracy across these runs to obtain a robust estimate of performance. For the BioHAR dataset, we adopt an 80/10/10 train/test/validation split, while the Sleep-EDF dataset reserves one subject for testing. For the UEA dataset, we follow the protocol established by TimesNet \cite{wu2023timesnet}, which provides only training and testing partitions, resulting in performance that corresponds to an upper-bound estimate, this is motivated by the fact that some datasets have only a few examples per class in the training data, and reserving part of it for validation could remove critical samples for those classes.

PRISM is evaluated against a comprehensive set of state-of-the-art baselines, including CNN-based models such as TSLANet \cite{eldele2024tslanet}, and TimesNet \cite{wu2023timesnet}; Transformer-based architectures including PatchTST \cite{yuqietal2023PatchTST} and iTransformer \cite{liu2023itransformer}; and simplified linear or MLP-based models such as DLinear \cite{zeng2023transformers}, LightTS \cite{zhang2022less}, and FiLM \cite{zhou2022film}. We also contextualise PRISM's performance against recent ultra-lightweight models and efficient architectures, including MiniRocket \cite{dempster2021minirocket}, LITE \cite{ismail2025look}, and a Mamba-based architecture \cite{gu2023mamba}, representing the current landscape of high-efficiency time-series classification.

To select Transformer based models' hyperparameters, we conducted a sweep over the Transformer depth (number of layers) and the number of attention heads on a representative subset of UEA datasets (following the TimesNet protocol) for the PatchTST architecture. For each (layers, heads) configuration, we computed the average test accuracy over this subset; the best-performing setting is obtained with 3 layers and 8 heads, which we use in our main experiments.

Finally, all baselines are implemented and evaluated within the same experimental framework, reusing shared components where possible and matching common hyperparameters across models whenever applicable to ensure a fair comparison.

\subsection{Results}

\begin{table*}[t]
\centering
\caption{UEA Results. \textbf{PRISM} is compared against 10 baselines, ordered from left to right by increasing parameter count (\textit{Avg Params}). 
\textcolor{blue}{\textbf{Blue}} indicates the highest mean accuracy for each dataset, while 
\textcolor[HTML]{D55E00}{\underline{Orange}} indicates the runner-up. 
Grey shading in the \textit{Avg Params}, \textit{Avg GFlops}, and \textit{Average} rows visualises relative magnitude (darker = larger). 
Results are reported as mean $\pm$ standard deviation over repeated runs.}

\label{tab:uea-results}
\resizebox{\textwidth}{!}{%
\begin{tabular}{lccccccccccc}
\toprule
 & \textbf{Ours} & \multicolumn{10}{c}{\textit{Baselines}} \\
\cmidrule(lr){2-2} \cmidrule(lr){3-12}
Dataset & \textbf{PRISM} & LITE & MiniROCKET & Mamba & iTransformer & PatchTST & TSLANet & FiLM & LightTS & DLinear & TimesNet \\
\midrule
\textit{Avg Params} & \cellcolor{gray!12}\textbf{\textit{66.26K}} & \cellcolor{gray!10}\textit{53.69K} & \cellcolor{gray!20}\textit{118.67K} & \cellcolor{gray!29}\textit{241.72K} & \cellcolor{gray!40}\textit{563.62K} & \cellcolor{gray!42}\textit{684.59K} & \cellcolor{gray!44}\textit{800.82K} & \cellcolor{gray!78}\textit{\textcolor{white}{10.93M}} & \cellcolor{gray!81}\textit{\textcolor{white}{13.46M}} & \cellcolor{gray!90}\textit{\textcolor{white}{26.69M}} & \cellcolor{gray!100}\textit{\textcolor{white}{56.59M}} \\
\textit{Avg GFlops} & \cellcolor{gray!17}\textbf{\textit{0.59G}} & \cellcolor{gray!26}\textit{1.57G} & \cellcolor{gray!44}\textit{9.34G} & \cellcolor{gray!35}\textit{3.82G} & \cellcolor{gray!10}\textit{0.28G} & \cellcolor{gray!38}\textit{5.45G} & \cellcolor{gray!35}\textit{3.78G} & \cellcolor{gray!28}\textit{1.79G} & \cellcolor{gray!34}\textit{3.33G} & \cellcolor{gray!40}\textit{6.36G} & \cellcolor{gray!100}\textbf{\textit{\textcolor{white}{3.05T}}} \\
\midrule
\midrule
ArticularyWordRecognition & 98.33 {\scriptsize$\pm$ 0.33} & \textcolor[HTML]{D55E00}{\underline{98.44 {\scriptsize$\pm$ 0.19}}} & \textcolor{blue}{\textbf{98.56 {\scriptsize$\pm$ 0.51}}} & 98.00 {\scriptsize$\pm$ 0.58} & \textcolor[HTML]{D55E00}{\underline{98.44 {\scriptsize$\pm$ 0.19}}} & 97.78 {\scriptsize$\pm$ 0.19} & 98.00 {\scriptsize$\pm$ 0.33} & 86.11 {\scriptsize$\pm$ 4.22} & 97.11 {\scriptsize$\pm$ 0.38} & 97.11 {\scriptsize$\pm$ 0.19} & 97.89 {\scriptsize$\pm$ 0.19} \\
AtrialFibrillation & \textcolor[HTML]{D55E00}{\underline{40.00 {\scriptsize$\pm$ 6.67}}} & 31.11 {\scriptsize$\pm$ 3.85} & 33.33 {\scriptsize$\pm$ 0.00} & 26.67 {\scriptsize$\pm$ 0.00} & 28.89 {\scriptsize$\pm$ 13.88} & 26.67 {\scriptsize$\pm$ 6.67} & 35.56 {\scriptsize$\pm$ 7.70} & 33.33 {\scriptsize$\pm$ 0.00} & \textcolor{blue}{\textbf{42.22 {\scriptsize$\pm$ 13.88}}} & 28.89 {\scriptsize$\pm$ 7.70} & 35.56 {\scriptsize$\pm$ 13.88} \\
BasicMotions & \textcolor{blue}{\textbf{\textbf{100.00 {\scriptsize$\pm$ 0.00}}}} & 75.00 {\scriptsize$\pm$ 43.30} & \textcolor{blue}{\textbf{100.00 {\scriptsize$\pm$ 0.00}}} & 94.17 {\scriptsize$\pm$ 10.10} & 90.00 {\scriptsize$\pm$ 2.50} & 60.83 {\scriptsize$\pm$ 3.82} & \textcolor{blue}{\textbf{100.00 {\scriptsize$\pm$ 0.00}}} & 37.50 {\scriptsize$\pm$ 19.53} & 94.17 {\scriptsize$\pm$ 8.04} & 85.00 {\scriptsize$\pm$ 2.50} & \textcolor[HTML]{D55E00}{\underline{99.17 {\scriptsize$\pm$ 1.44}}} \\
Cricket & \textcolor[HTML]{D55E00}{\underline{97.69 {\scriptsize$\pm$ 0.80}}} & \textcolor{blue}{\textbf{98.61 {\scriptsize$\pm$ 0.00}}} & \textcolor{blue}{\textbf{98.61 {\scriptsize$\pm$ 0.00}}} & \textcolor{blue}{\textbf{98.61 {\scriptsize$\pm$ 0.00}}} & 87.04 {\scriptsize$\pm$ 0.80} & 91.67 {\scriptsize$\pm$ 2.41} & \textcolor[HTML]{D55E00}{\underline{97.69 {\scriptsize$\pm$ 0.80}}} & 53.24 {\scriptsize$\pm$ 8.93} & 89.35 {\scriptsize$\pm$ 0.80} & 91.20 {\scriptsize$\pm$ 0.80} & 91.20 {\scriptsize$\pm$ 3.50} \\
DuckDuckGeese & 37.33 {\scriptsize$\pm$ 11.02} & 54.00 {\scriptsize$\pm$ 5.29} & 52.67 {\scriptsize$\pm$ 2.31} & \textcolor[HTML]{D55E00}{\underline{58.00 {\scriptsize$\pm$ 7.21}}} & 42.00 {\scriptsize$\pm$ 2.00} & 17.33 {\scriptsize$\pm$ 3.06} & \textcolor{blue}{\textbf{62.67 {\scriptsize$\pm$ 4.16}}} & 20.00 {\scriptsize$\pm$ 0.00} & 31.33 {\scriptsize$\pm$ 16.29} & 38.67 {\scriptsize$\pm$ 12.22} & 56.67 {\scriptsize$\pm$ 4.62} \\
ERing & 80.37 {\scriptsize$\pm$ 1.34} & 16.67 {\scriptsize$\pm$ 0.00} & \textcolor{blue}{\textbf{97.78 {\scriptsize$\pm$ 0.37}}} & 92.96 {\scriptsize$\pm$ 1.34} & 91.36 {\scriptsize$\pm$ 1.19} & \textcolor[HTML]{D55E00}{\underline{94.57 {\scriptsize$\pm$ 1.19}}} & 87.78 {\scriptsize$\pm$ 0.74} & 77.53 {\scriptsize$\pm$ 0.77} & 90.00 {\scriptsize$\pm$ 1.34} & 90.25 {\scriptsize$\pm$ 0.21} & 93.70 {\scriptsize$\pm$ 0.37} \\
EigenWorms & 62.60 {\scriptsize$\pm$ 3.82} & \textcolor[HTML]{D55E00}{\underline{71.76 {\scriptsize$\pm$ 16.74}}} & 48.85 {\scriptsize$\pm$ 2.29} & 52.84 {\scriptsize$\pm$ 8.50} & 47.07 {\scriptsize$\pm$ 1.92} & 31.04 {\scriptsize$\pm$ 10.70} & \textcolor{blue}{\textbf{77.35 {\scriptsize$\pm$ 1.59}}} & 31.81 {\scriptsize$\pm$ 15.69} & 48.09 {\scriptsize$\pm$ 1.53} & 33.08 {\scriptsize$\pm$ 1.59} & 35.11 {\scriptsize$\pm$ 14.50} \\
Epilepsy & 89.61 {\scriptsize$\pm$ 1.51} & \textcolor[HTML]{D55E00}{\underline{96.86 {\scriptsize$\pm$ 0.84}}} & \textcolor{blue}{\textbf{100.00 {\scriptsize$\pm$ 0.00}}} & 96.38 {\scriptsize$\pm$ 0.00} & 71.26 {\scriptsize$\pm$ 1.67} & 96.14 {\scriptsize$\pm$ 1.11} & 93.48 {\scriptsize$\pm$ 2.17} & 56.28 {\scriptsize$\pm$ 8.14} & 85.75 {\scriptsize$\pm$ 9.06} & 45.17 {\scriptsize$\pm$ 1.51} & 86.23 {\scriptsize$\pm$ 1.45} \\
EthanolConcentration & 24.21 {\scriptsize$\pm$ 1.95} & 25.10 {\scriptsize$\pm$ 0.38} & 23.57 {\scriptsize$\pm$ 4.23} & 23.45 {\scriptsize$\pm$ 1.80} & 23.45 {\scriptsize$\pm$ 1.16} & 24.33 {\scriptsize$\pm$ 1.01} & 24.84 {\scriptsize$\pm$ 0.44} & \textcolor{blue}{\textbf{26.87 {\scriptsize$\pm$ 3.53}}} & \textcolor[HTML]{D55E00}{\underline{26.36 {\scriptsize$\pm$ 1.88}}} & 25.98 {\scriptsize$\pm$ 3.32} & 25.98 {\scriptsize$\pm$ 1.22} \\
FaceDetection & 62.68 {\scriptsize$\pm$ 0.72} & \textcolor[HTML]{D55E00}{\underline{66.80 {\scriptsize$\pm$ 1.12}}} & 59.90 {\scriptsize$\pm$ 0.80} & 65.35 {\scriptsize$\pm$ 1.81} & 64.77 {\scriptsize$\pm$ 1.76} & 64.78 {\scriptsize$\pm$ 0.48} & 57.04 {\scriptsize$\pm$ 0.60} & 62.93 {\scriptsize$\pm$ 1.88} & \textcolor{blue}{\textbf{66.85 {\scriptsize$\pm$ 0.28}}} & 66.39 {\scriptsize$\pm$ 1.54} & 65.22 {\scriptsize$\pm$ 0.73} \\
FingerMovements & \textcolor[HTML]{D55E00}{\underline{57.33 {\scriptsize$\pm$ 4.04}}} & 52.33 {\scriptsize$\pm$ 4.16} & \textcolor{blue}{\textbf{60.00 {\scriptsize$\pm$ 6.24}}} & 54.33 {\scriptsize$\pm$ 4.73} & 54.00 {\scriptsize$\pm$ 3.61} & 52.67 {\scriptsize$\pm$ 3.06} & 54.67 {\scriptsize$\pm$ 2.31} & 49.00 {\scriptsize$\pm$ 1.00} & 50.33 {\scriptsize$\pm$ 8.74} & 48.67 {\scriptsize$\pm$ 3.79} & 50.67 {\scriptsize$\pm$ 2.08} \\
HandMovementDirection & 32.88 {\scriptsize$\pm$ 5.12} & 29.28 {\scriptsize$\pm$ 7.92} & 36.49 {\scriptsize$\pm$ 3.58} & \textcolor[HTML]{D55E00}{\underline{54.50 {\scriptsize$\pm$ 14.82}}} & 51.35 {\scriptsize$\pm$ 2.70} & 44.14 {\scriptsize$\pm$ 2.81} & 31.08 {\scriptsize$\pm$ 7.02} & 33.33 {\scriptsize$\pm$ 2.81} & 53.60 {\scriptsize$\pm$ 2.06} & 53.60 {\scriptsize$\pm$ 2.06} & \textcolor{blue}{\textbf{54.95 {\scriptsize$\pm$ 2.06}}} \\
Handwriting & 18.04 {\scriptsize$\pm$ 5.10} & 19.65 {\scriptsize$\pm$ 26.80} & \textcolor{blue}{\textbf{45.37 {\scriptsize$\pm$ 3.04}}} & 8.43 {\scriptsize$\pm$ 7.68} & 23.10 {\scriptsize$\pm$ 0.48} & 14.08 {\scriptsize$\pm$ 2.11} & \textcolor[HTML]{D55E00}{\underline{33.37 {\scriptsize$\pm$ 0.59}}} & 6.59 {\scriptsize$\pm$ 2.58} & 13.18 {\scriptsize$\pm$ 1.13} & 12.94 {\scriptsize$\pm$ 0.12} & 23.49 {\scriptsize$\pm$ 3.15} \\
Heartbeat & \textcolor{blue}{\textbf{\textbf{80.00 {\scriptsize$\pm$ 1.76}}}} & 74.15 {\scriptsize$\pm$ 1.69} & 75.45 {\scriptsize$\pm$ 1.85} & \textcolor[HTML]{D55E00}{\underline{76.59 {\scriptsize$\pm$ 0.98}}} & 74.47 {\scriptsize$\pm$ 1.71} & 59.67 {\scriptsize$\pm$ 12.93} & 73.82 {\scriptsize$\pm$ 1.02} & 72.20 {\scriptsize$\pm$ 0.00} & 74.63 {\scriptsize$\pm$ 0.98} & 72.52 {\scriptsize$\pm$ 2.46} & 74.96 {\scriptsize$\pm$ 1.97} \\
InsectWingbeat & \textcolor[HTML]{D55E00}{\underline{70.48 {\scriptsize$\pm$ 0.30}}} & 61.69 {\scriptsize$\pm$ 0.37} & 62.32 {\scriptsize$\pm$ 0.49} & 63.25 {\scriptsize$\pm$ 0.84} & \textcolor{blue}{\textbf{71.98 {\scriptsize$\pm$ 0.14}}} & 59.17 {\scriptsize$\pm$ 0.32} & 54.54 {\scriptsize$\pm$ 0.50} & 67.51 {\scriptsize$\pm$ 0.28} & 70.27 {\scriptsize$\pm$ 0.25} & 15.78 {\scriptsize$\pm$ 1.90} & 61.96 {\scriptsize$\pm$ 0.47} \\
JapaneseVowels & 97.03 {\scriptsize$\pm$ 0.54} & \textcolor[HTML]{D55E00}{\underline{97.93 {\scriptsize$\pm$ 0.16}}} & \textcolor{blue}{\textbf{98.02 {\scriptsize$\pm$ 0.41}}} & 96.58 {\scriptsize$\pm$ 0.56} & 96.67 {\scriptsize$\pm$ 0.31} & 94.95 {\scriptsize$\pm$ 0.41} & 96.94 {\scriptsize$\pm$ 0.41} & 89.37 {\scriptsize$\pm$ 3.04} & 94.86 {\scriptsize$\pm$ 0.54} & 96.31 {\scriptsize$\pm$ 0.41} & 97.84 {\scriptsize$\pm$ 0.54} \\
LSST & 31.74 {\scriptsize$\pm$ 0.12} & 31.81 {\scriptsize$\pm$ 0.17} & \textcolor{blue}{\textbf{42.50 {\scriptsize$\pm$ 5.95}}} & 26.16 {\scriptsize$\pm$ 9.20} & 33.67 {\scriptsize$\pm$ 0.54} & \textcolor[HTML]{D55E00}{\underline{41.73 {\scriptsize$\pm$ 5.53}}} & 33.82 {\scriptsize$\pm$ 1.76} & 33.97 {\scriptsize$\pm$ 0.63} & 35.09 {\scriptsize$\pm$ 2.40} & 31.58 {\scriptsize$\pm$ 0.05} & 34.12 {\scriptsize$\pm$ 1.77} \\
Libras & 85.56 {\scriptsize$\pm$ 2.42} & 88.52 {\scriptsize$\pm$ 2.57} & \textcolor{blue}{\textbf{93.70 {\scriptsize$\pm$ 0.85}}} & 84.81 {\scriptsize$\pm$ 6.72} & \textcolor[HTML]{D55E00}{\underline{88.70 {\scriptsize$\pm$ 0.64}}} & 74.81 {\scriptsize$\pm$ 1.16} & 85.00 {\scriptsize$\pm$ 0.96} & 78.33 {\scriptsize$\pm$ 2.42} & 70.93 {\scriptsize$\pm$ 8.10} & 67.96 {\scriptsize$\pm$ 1.40} & 77.78 {\scriptsize$\pm$ 2.00} \\
MotorImagery & 48.00 {\scriptsize$\pm$ 4.36} & 48.67 {\scriptsize$\pm$ 2.31} & \textcolor{blue}{\textbf{58.33 {\scriptsize$\pm$ 4.16}}} & 50.00 {\scriptsize$\pm$ 1.00} & \textcolor[HTML]{D55E00}{\underline{55.00 {\scriptsize$\pm$ 3.46}}} & 52.67 {\scriptsize$\pm$ 3.79} & 52.00 {\scriptsize$\pm$ 6.08} & 50.00 {\scriptsize$\pm$ 0.00} & 51.67 {\scriptsize$\pm$ 4.93} & \textcolor[HTML]{D55E00}{\underline{55.00 {\scriptsize$\pm$ 4.58}}} & 48.00 {\scriptsize$\pm$ 2.65} \\
NATOPS & 91.67 {\scriptsize$\pm$ 2.55} & \textcolor{blue}{\textbf{95.74 {\scriptsize$\pm$ 1.16}}} & 91.67 {\scriptsize$\pm$ 0.00} & 92.41 {\scriptsize$\pm$ 2.51} & 83.52 {\scriptsize$\pm$ 1.28} & 75.00 {\scriptsize$\pm$ 1.47} & \textcolor[HTML]{D55E00}{\underline{94.07 {\scriptsize$\pm$ 2.25}}} & 78.70 {\scriptsize$\pm$ 1.70} & 93.33 {\scriptsize$\pm$ 2.22} & 93.70 {\scriptsize$\pm$ 0.32} & 90.74 {\scriptsize$\pm$ 3.70} \\
PEMS-SF & \textcolor[HTML]{D55E00}{\underline{81.70 {\scriptsize$\pm$ 0.67}}} & 77.07 {\scriptsize$\pm$ 3.18} & \textcolor{blue}{\textbf{84.20 {\scriptsize$\pm$ 2.61}}} & 73.99 {\scriptsize$\pm$ 1.00} & 80.15 {\scriptsize$\pm$ 1.20} & 67.44 {\scriptsize$\pm$ 19.36} & 74.18 {\scriptsize$\pm$ 1.20} & 66.09 {\scriptsize$\pm$ 7.54} & 80.15 {\scriptsize$\pm$ 5.90} & 69.56 {\scriptsize$\pm$ 4.41} & 81.31 {\scriptsize$\pm$ 2.34} \\
PenDigits & 97.44 {\scriptsize$\pm$ 0.32} & \textcolor[HTML]{D55E00}{\underline{98.36 {\scriptsize$\pm$ 0.29}}} & 10.38 {\scriptsize$\pm$ 0.00} & \textcolor{blue}{\textbf{98.39 {\scriptsize$\pm$ 0.54}}} & 96.86 {\scriptsize$\pm$ 0.26} & 94.35 {\scriptsize$\pm$ 1.07} & 98.02 {\scriptsize$\pm$ 0.26} & 90.32 {\scriptsize$\pm$ 2.54} & 94.42 {\scriptsize$\pm$ 1.96} & 90.17 {\scriptsize$\pm$ 0.15} & 97.24 {\scriptsize$\pm$ 0.44} \\
PhonemeSpectra & \textcolor[HTML]{D55E00}{\underline{25.81 {\scriptsize$\pm$ 1.03}}} & 25.03 {\scriptsize$\pm$ 0.91} & 17.91 {\scriptsize$\pm$ 1.04} & \textcolor{blue}{\textbf{26.27 {\scriptsize$\pm$ 0.85}}} & 6.66 {\scriptsize$\pm$ 0.48} & 7.22 {\scriptsize$\pm$ 2.21} & 21.83 {\scriptsize$\pm$ 0.72} & 7.87 {\scriptsize$\pm$ 0.44} & 5.68 {\scriptsize$\pm$ 0.31} & 6.33 {\scriptsize$\pm$ 0.59} & 10.76 {\scriptsize$\pm$ 0.58} \\
RacketSports & 73.90 {\scriptsize$\pm$ 1.37} & 87.72 {\scriptsize$\pm$ 1.00} & \textcolor{blue}{\textbf{90.57 {\scriptsize$\pm$ 0.76}}} & \textcolor[HTML]{D55E00}{\underline{89.47 {\scriptsize$\pm$ 5.70}}} & 76.32 {\scriptsize$\pm$ 2.87} & 72.59 {\scriptsize$\pm$ 0.38} & 77.19 {\scriptsize$\pm$ 0.38} & 58.99 {\scriptsize$\pm$ 6.11} & 69.52 {\scriptsize$\pm$ 2.11} & 75.66 {\scriptsize$\pm$ 1.74} & 86.62 {\scriptsize$\pm$ 2.11} \\
SelfRegulationSCP1 & 86.69 {\scriptsize$\pm$ 0.90} & 85.21 {\scriptsize$\pm$ 1.20} & \textcolor[HTML]{D55E00}{\underline{91.01 {\scriptsize$\pm$ 0.52}}} & 87.14 {\scriptsize$\pm$ 1.20} & 89.65 {\scriptsize$\pm$ 1.10} & 81.80 {\scriptsize$\pm$ 2.90} & 87.71 {\scriptsize$\pm$ 1.81} & 90.10 {\scriptsize$\pm$ 1.81} & \textcolor[HTML]{D55E00}{\underline{91.01 {\scriptsize$\pm$ 1.38}}} & \textcolor{blue}{\textbf{91.81 {\scriptsize$\pm$ 0.34}}} & 81.91 {\scriptsize$\pm$ 6.93} \\
SelfRegulationSCP2 & \textcolor[HTML]{D55E00}{\underline{54.81 {\scriptsize$\pm$ 0.85}}} & 51.48 {\scriptsize$\pm$ 2.57} & 51.85 {\scriptsize$\pm$ 1.70} & 53.89 {\scriptsize$\pm$ 2.55} & \textcolor[HTML]{D55E00}{\underline{54.81 {\scriptsize$\pm$ 3.06}}} & 52.78 {\scriptsize$\pm$ 2.00} & \textcolor{blue}{\textbf{55.19 {\scriptsize$\pm$ 1.16}}} & 49.81 {\scriptsize$\pm$ 0.32} & 53.33 {\scriptsize$\pm$ 6.11} & 49.26 {\scriptsize$\pm$ 4.72} & 52.41 {\scriptsize$\pm$ 5.45} \\
SpokenArabicDigits & 98.56 {\scriptsize$\pm$ 0.14} & \textcolor{blue}{\textbf{99.59 {\scriptsize$\pm$ 0.23}}} & 98.12 {\scriptsize$\pm$ 0.18} & \textcolor[HTML]{D55E00}{\underline{98.95 {\scriptsize$\pm$ 0.28}}} & 98.14 {\scriptsize$\pm$ 0.52} & 96.33 {\scriptsize$\pm$ 1.61} & 98.44 {\scriptsize$\pm$ 0.42} & 96.95 {\scriptsize$\pm$ 0.20} & 97.38 {\scriptsize$\pm$ 0.25} & 96.45 {\scriptsize$\pm$ 0.34} & 97.54 {\scriptsize$\pm$ 0.32} \\
StandWalkJump & 40.00 {\scriptsize$\pm$ 6.67} & 33.33 {\scriptsize$\pm$ 0.00} & 35.56 {\scriptsize$\pm$ 3.85} & 26.67 {\scriptsize$\pm$ 11.55} & 33.33 {\scriptsize$\pm$ 17.64} & \textcolor{blue}{\textbf{51.11 {\scriptsize$\pm$ 3.85}}} & 28.89 {\scriptsize$\pm$ 3.85} & 37.78 {\scriptsize$\pm$ 3.85} & \textcolor[HTML]{D55E00}{\underline{46.67 {\scriptsize$\pm$ 6.67}}} & 31.11 {\scriptsize$\pm$ 13.88} & 40.00 {\scriptsize$\pm$ 6.67} \\
UWaveGestureLibrary & 85.62 {\scriptsize$\pm$ 2.19} & \textcolor[HTML]{D55E00}{\underline{90.62 {\scriptsize$\pm$ 0.31}}} & \textcolor{blue}{\textbf{92.19 {\scriptsize$\pm$ 0.31}}} & 85.73 {\scriptsize$\pm$ 4.42} & 85.94 {\scriptsize$\pm$ 1.36} & 84.27 {\scriptsize$\pm$ 0.36} & \textcolor[HTML]{D55E00}{\underline{90.62 {\scriptsize$\pm$ 0.31}}} & 78.44 {\scriptsize$\pm$ 4.33} & 82.81 {\scriptsize$\pm$ 0.62} & 81.46 {\scriptsize$\pm$ 0.48} & 86.46 {\scriptsize$\pm$ 1.26} \\
\midrule
Average & \cellcolor{gray!56}\textbf{67.24} & \cellcolor{gray!48}64.91 & \cellcolor{gray!56}67.20 & \cellcolor{gray!57}67.38 & \cellcolor{gray!50}65.47 & \cellcolor{gray!35}61.45 & \cellcolor{gray!60}68.13 & \cellcolor{gray!15}55.90 & \cellcolor{gray!50}65.52 & \cellcolor{gray!30}60.06 & \cellcolor{gray!54}66.74 \\
\bottomrule
\end{tabular}
}%
\end{table*}

\begin{table*}[t]
\centering
\caption{Human Activity Recognition and Biomedical Signal Results. \textbf{PRISM} is compared against 10 baselines, ordered from left to right by increasing parameter count (\textit{Avg Params}). 
\textcolor{blue}{\textbf{Best}} indicates the highest mean accuracy for each dataset, while 
\textcolor[HTML]{D55E00}{\underline{Orange}} indicates the runner-up. 
Grey shading in the \textit{Avg Params}, \textit{Avg GFlops}, and \textit{Average} rows visualises relative magnitude (darker = larger). 
Results are reported as mean $\pm$ standard deviation over repeated runs.}
\label{tab:human-activity-recognition-and-biomedical-signal-results}
\resizebox{\textwidth}{!}{%
\begin{tabular}{lccccccccccc}
\toprule
 & \textbf{Ours} & \multicolumn{10}{c}{\textit{Baselines}} \\
\cmidrule(lr){2-2} \cmidrule(lr){3-12}
Dataset & \textbf{PRISM} & LITE & MiniROCKET & Mamba & PatchTST & iTransformer & TSLANet & LightTS & DLinear & FiLM & TimesNet \\
\midrule
\textit{Avg Params} & \cellcolor{gray!10}\textbf{\textit{13.59K}} & \cellcolor{gray!12}\textit{16.84K} & \cellcolor{gray!28}\textit{72.55K} & \cellcolor{gray!40}\textit{235.81K} & \cellcolor{gray!48}\textit{485.79K} & \cellcolor{gray!48}\textit{491.73K} & \cellcolor{gray!53}\textit{762.32K} & \cellcolor{gray!63}\textit{\textcolor{white}{1.85M}} & \cellcolor{gray!70}\textit{\textcolor{white}{3.64M}} & \cellcolor{gray!83}\textit{\textcolor{white}{12.59M}} & \cellcolor{gray!100}\textit{\textcolor{white}{56.69M}} \\
\textit{Avg GFlops} & \cellcolor{gray!10}\textbf{\textit{0.04G}} & \cellcolor{gray!24}\textit{0.27G} & \cellcolor{gray!30}\textit{0.55G} & \cellcolor{gray!47}\textit{5.11G} & \cellcolor{gray!38}\textit{1.63G} & \cellcolor{gray!11}\textit{0.05G} & \cellcolor{gray!40}\textit{2.12G} & \cellcolor{gray!13}\textit{0.07G} & \cellcolor{gray!18}\textit{0.12G} & \cellcolor{gray!30}\textit{0.58G} & \cellcolor{gray!100}\textbf{\textit{\textcolor{white}{4.07T}}} \\
\addlinespace[1ex]
\midrule
\multicolumn{12}{c}{\textbf{Human Activity Recognition (HAR)}} \\
\midrule
HHAR\_SA & 97.48 {\scriptsize$\pm$ 0.22} & \textcolor{blue}{\textbf{98.16 {\scriptsize$\pm$ 0.18}}} & 95.90 {\scriptsize$\pm$ 1.10} & \textcolor[HTML]{D55E00}{\underline{97.72 {\scriptsize$\pm$ 0.26}}} & 85.35 {\scriptsize$\pm$ 2.60} & 91.96 {\scriptsize$\pm$ 0.49} & 96.60 {\scriptsize$\pm$ 0.94} & 93.58 {\scriptsize$\pm$ 0.73} & 43.43 {\scriptsize$\pm$ 2.61} & 86.99 {\scriptsize$\pm$ 1.24} & 95.38 {\scriptsize$\pm$ 0.58} \\
UCIHAR & \textcolor{blue}{\textbf{96.37 {\scriptsize$\pm$ 1.65}}} & 92.55 {\scriptsize$\pm$ 0.39} & \textcolor[HTML]{D55E00}{\underline{95.92 {\scriptsize$\pm$ 0.86}}} & 91.23 {\scriptsize$\pm$ 0.86} & 79.21 {\scriptsize$\pm$ 2.50} & 90.88 {\scriptsize$\pm$ 0.69} & 92.53 {\scriptsize$\pm$ 1.50} & 88.96 {\scriptsize$\pm$ 1.93} & 51.87 {\scriptsize$\pm$ 2.34} & 87.68 {\scriptsize$\pm$ 1.12} & 89.02 {\scriptsize$\pm$ 1.34} \\
WISDM & 94.65 {\scriptsize$\pm$ 0.65} & \textcolor[HTML]{D55E00}{\underline{97.79 {\scriptsize$\pm$ 0.39}}} & 95.20 {\scriptsize$\pm$ 0.90} & \textcolor{blue}{\textbf{97.88 {\scriptsize$\pm$ 0.53}}} & 85.49 {\scriptsize$\pm$ 2.36} & 83.89 {\scriptsize$\pm$ 0.76} & 95.38 {\scriptsize$\pm$ 0.54} & 84.95 {\scriptsize$\pm$ 0.45} & 60.41 {\scriptsize$\pm$ 0.14} & 80.06 {\scriptsize$\pm$ 1.67} & 87.24 {\scriptsize$\pm$ 4.87} \\
\addlinespace
\textit{HAR Average} & \cellcolor{gray!60}\textbf{96.16} & \cellcolor{gray!59}96.16 & \cellcolor{gray!59}95.67 & \cellcolor{gray!59}95.61 & \cellcolor{gray!46}83.35 & \cellcolor{gray!52}88.91 & \cellcolor{gray!58}94.84 & \cellcolor{gray!52}89.16 & \cellcolor{gray!15}51.90 & \cellcolor{gray!48}84.91 & \cellcolor{gray!54}90.55 \\
\addlinespace[1ex]
\midrule
\multicolumn{12}{c}{\textbf{Biomedical Signals}} \\
\midrule
ECG & 97.76 {\scriptsize$\pm$ 0.24} & \textcolor[HTML]{D55E00}{\underline{98.40 {\scriptsize$\pm$ 0.09}}} & 97.76 {\scriptsize$\pm$ 0.24} & 98.27 {\scriptsize$\pm$ 0.08} & 93.21 {\scriptsize$\pm$ 0.33} & 96.89 {\scriptsize$\pm$ 0.08} & \textcolor{blue}{\textbf{98.51 {\scriptsize$\pm$ 0.11}}} & 97.51 {\scriptsize$\pm$ 0.20} & 90.41 {\scriptsize$\pm$ 0.13} & 97.63 {\scriptsize$\pm$ 0.03} & 98.21 {\scriptsize$\pm$ 0.14} \\
Sleep\_EDF & \textcolor[HTML]{D55E00}{\underline{85.02 {\scriptsize$\pm$ 0.34}}} & \textcolor{blue}{\textbf{85.30 {\scriptsize$\pm$ 0.37}}} & 80.55 {\scriptsize$\pm$ 2.72} & 83.75 {\scriptsize$\pm$ 1.40} & 77.66 {\scriptsize$\pm$ 1.41} & 39.21 {\scriptsize$\pm$ 0.53} & 83.67 {\scriptsize$\pm$ 0.92} & 57.60 {\scriptsize$\pm$ 7.51} & 23.92 {\scriptsize$\pm$ 2.03} & 49.66 {\scriptsize$\pm$ 1.89} & 60.18 {\scriptsize$\pm$ 19.85} \\
\addlinespace
\textit{Bio Average} & \cellcolor{gray!59}\textbf{91.39} & \cellcolor{gray!60}91.85 & \cellcolor{gray!56}89.16 & \cellcolor{gray!58}91.01 & \cellcolor{gray!51}85.43 & \cellcolor{gray!29}68.05 & \cellcolor{gray!59}91.09 & \cellcolor{gray!41}77.55 & \cellcolor{gray!15}57.16 & \cellcolor{gray!36}73.64 & \cellcolor{gray!43}79.20 \\
\addlinespace[1ex]
\midrule
\textbf{Overall Average} & \cellcolor{gray!59}\textbf{94.25} & \cellcolor{gray!60}94.44 & \cellcolor{gray!58}93.07 & \cellcolor{gray!59}93.77 & \cellcolor{gray!48}84.18 & \cellcolor{gray!44}80.57 & \cellcolor{gray!58}93.34 & \cellcolor{gray!48}84.52 & \cellcolor{gray!15}54.01 & \cellcolor{gray!44}80.40 & \cellcolor{gray!50}86.01 \\
\bottomrule
\end{tabular}
}%
\end{table*}

To assess how effectively PRISM navigates the trade-off between resource efficiency and representation power, Tables \ref{tab:uea-results} and \ref{tab:human-activity-recognition-and-biomedical-signal-results} report the classification accuracy alongside the computational demands (Avg Params and Avg GFlops) for the UEA and Bio/HAR benchmarks, respectively. Notably, to prevent disproportionate skewing of the computational metrics, the two highest-dimensional datasets, DuckDuckGeese (1,345 channels) and PEMS-SF (963 channels), were excluded from the Avg Params and Avg GFlops calculations. Because their extreme dimensionality constitutes a significant outlier with respect to the rest of the archive, their inclusion would misrepresent the model's typical hardware requirements. These datasets are, however, retained in the calculation of the overall average accuracy.
Focusing on the UEA archive, PRISM achieves a strong mean accuracy of 67.24\%, placing it consistently ahead of MLP-based baselines (DLinear, FiLM, LightTS) while rivaling more complex convolutional architectures such as TimesNet. Notably, PRISM outperforms the feature-based MiniROCKET (67.20\%) and its closest lightweight competitor, LITE (64.91\%), while remaining highly competitive with the state-space model Mamba (67.38\%). Demonstrating its capacity as an effective feature extractor, PRISM excels even on highly complex tasks such as \textsc{PhonemeSpectra}, a 39--class audio benchmark where most baselines struggle to exceed 10\% accuracy but in which PRISM reaches 25.81\%, closely following Mamba (26.27\%) while requiring $5\times$ fewer parameters.

On datasets where discriminative cues are strongly localised, including \textsc{BasicMotions}, \textsc{PenDigits}, and \textsc{SpokenArabicDigits}, PRISM reaches accuracy levels nearly identical to deeper CNNs and transformer--based approaches. This suggests that for problems dominated by local temporal motifs, architectural complexity has limited impact on achievable performance. However, due to its channel-independent design, PRISM falls behind in datasets where complex cross-channel or spatial dependencies are vital. Specific examples include \textsc{MotorImagery} (a 64-channel EEG dataset) and \textsc{HandMovementDirection} (a 10-channel MEG dataset), where PRISM reaches accuracies of 48.00\% and 32.88\%, respectively. In contrast, alternative architectures like MiniROCKET achieve stronger results on these datasets (58.33\% and 36.49\%, respectively). Furthermore, models such as Mamba, which employ early cross-channel interactions, are inherently better suited to capture these spatial structures, outperforming PRISM with an accuracy of 54.50\% on \textsc{HandMovementDirection}. We note this as a limitation, highlighting that architectures explicitly designed to model inter-channel relationships remain advantageous in such contexts; however, this often comes at the cost of increased complexity and the risk of creating spurious correlations between channels.

Our results align with the limitations of the UEA archive discussed in Section \ref{subsec:datasets}: the difference between the best-performing model and the simplest baseline (DLinear) on these benchmarks is within 8\%, indicating that more sophisticated deep learning architectures do not necessarily translate into substantially higher accuracy on this specific archive.

Table \ref{tab:human-activity-recognition-and-biomedical-signal-results} reports the results on the biomedical and HAR datasets. In this setting, LITE (94.44\% overall average) slightly outperforms PRISM (94.25\%), while PRISM outperforms Mamba (93.77\%) in terms of absolute mean accuracy. In contrast to the UEA benchmarks, the simple linear baseline DLinear performs poorly here, achieving only 51.90\% and 57.16\% average accuracy respectively, underscoring the greater complexity and richer temporal structure present in these datasets. 

Notably, PRISM delivers particularly strong results on the \textsc{Sleep EDF} dataset, achieving 85.02\%. This is highly competitive with the top-performing model, LITE (85.30\%), and outperforms TSLANet (83.67\%), while requiring approximately $8.5\times$ and $75\times$ lower GFLOPS and $2.3\times$ and $178\times$ fewer parameters, respectively, suggesting that its architecture is able to handle noisy electrophysiological recordings efficiently. The transformer-based models, on the other hand, lag behind on this benchmark, for instance, iTransformer only reaches 39.21\%, indicating ongoing challenges in effectively modelling long input lengths without degradation in representation quality.

\begin{figure*}[t] 
    \centering
    \includegraphics[width=\textwidth]{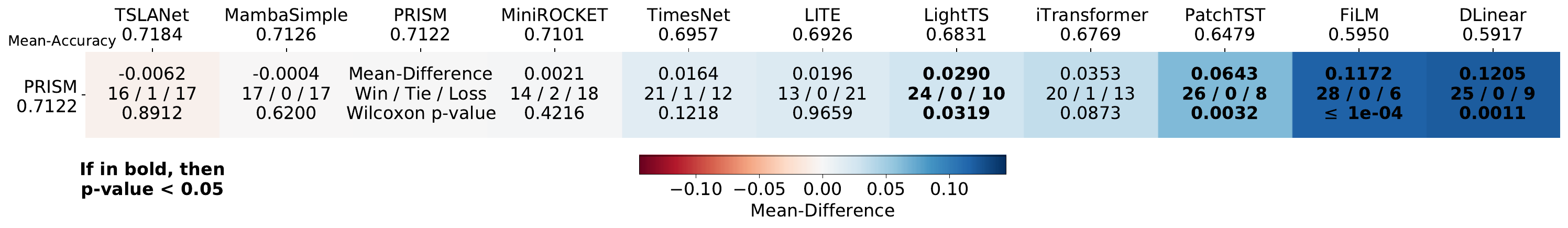} 
    \caption{Statistical comparison of model performance across the combined UEA and Bio/HAR datasets. The matrix displays the Wilcoxon signed-rank test $p$-values, mean differences, and how many times each model wins, ties, or loses against PRISM. Bold values indicate statistical significance ($p < 0.05$).}
    \label{fig:wilcoxon_matrix} 
\end{figure*}

Figure \ref{fig:wilcoxon_matrix} summarises both Tables \ref{tab:uea-results} and \ref{tab:human-activity-recognition-and-biomedical-signal-results} in a Multi-Comparison Matrix (MCM) \cite{ismail2023approach} (as used in recent works such as LITE).
PRISM achieves a mean accuracy of 71.22\%, closely following the top-performing models TSLANet (71.84\%) and Mamba (71.26\%) while requiring approximately $6\times$ lower FLOPs and $15\times$ and $6\times$ fewer parameters, respectively. Furthermore, the Wilcoxon signed-rank test confirms that there is no statistically significant difference in performance between PRISM and these leading architectures, confirming statistical parity. While PRISM outperforms several competitive models in terms of absolute mean accuracy, including MiniROCKET (71.01\%), TimesNet (69.57\%), and LITE (69.26\%), the test indicates these differences are also not statistically significant. Similarly, while PRISM achieves a higher absolute accuracy than iTransformer (67.69\%), this outperformance is not statistically significant. This gap is heavily influenced by iTransformer's poor performance on the Bio/HAR datasets, where it only averages around 81\%. Crucially, PRISM demonstrates a clear, statistically significant advantage over the remaining baselines, outperforming models such as LightTS, PatchTST, FiLM, and DLinear.

\subsection{Complexity Analysis}

\begin{figure*}[t]
  \centering
  \begin{subfigure}{0.48\textwidth}
    \centering
    \includegraphics[width=\linewidth]{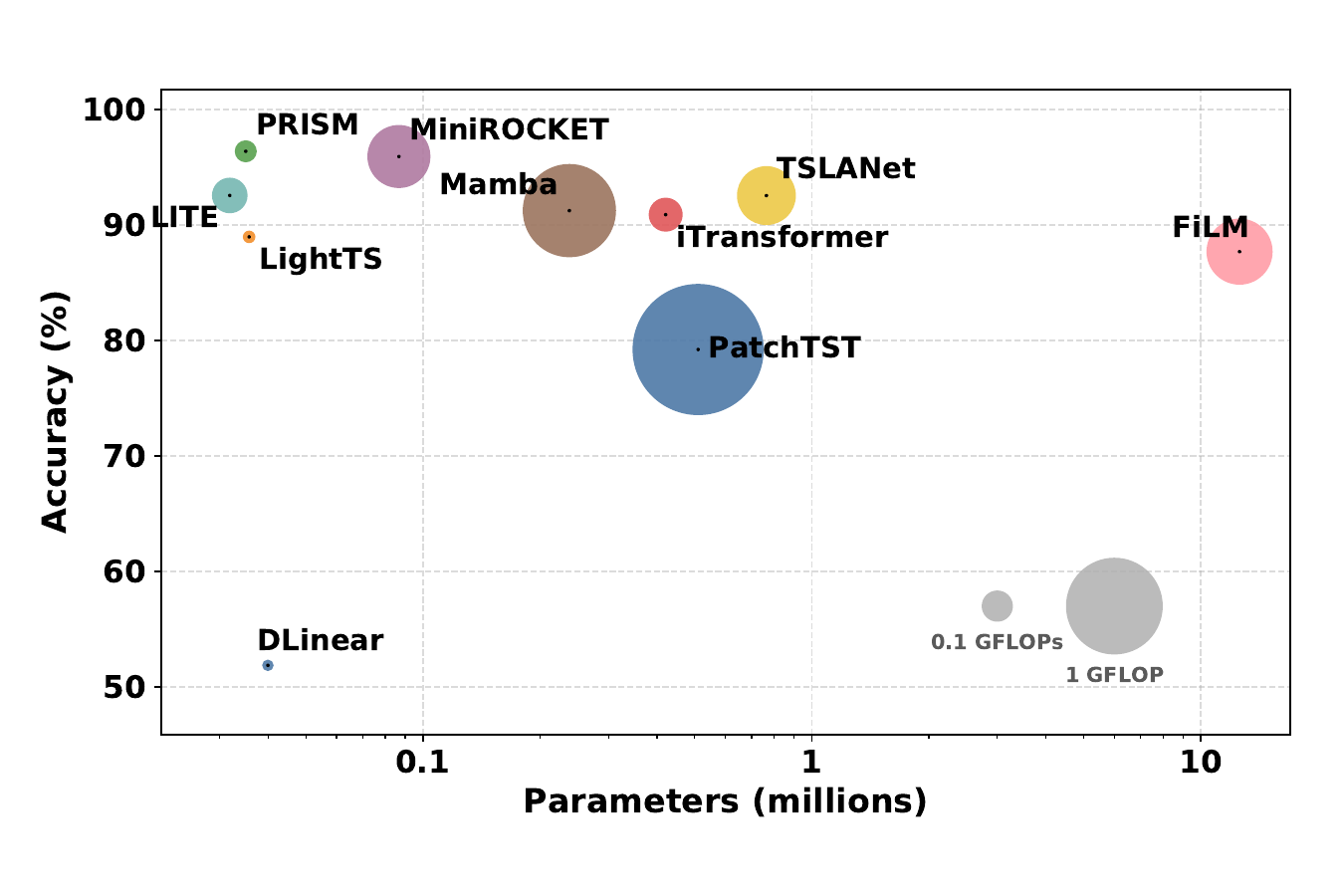}
    \caption{UCI-HAR — accuracy vs. parameters and complexity }
    \label{fig:accuracy_vs_gflops_ucihar}
  \end{subfigure}\hfill
  \begin{subfigure}{0.48\textwidth}
    \centering
    \includegraphics[width=\linewidth]{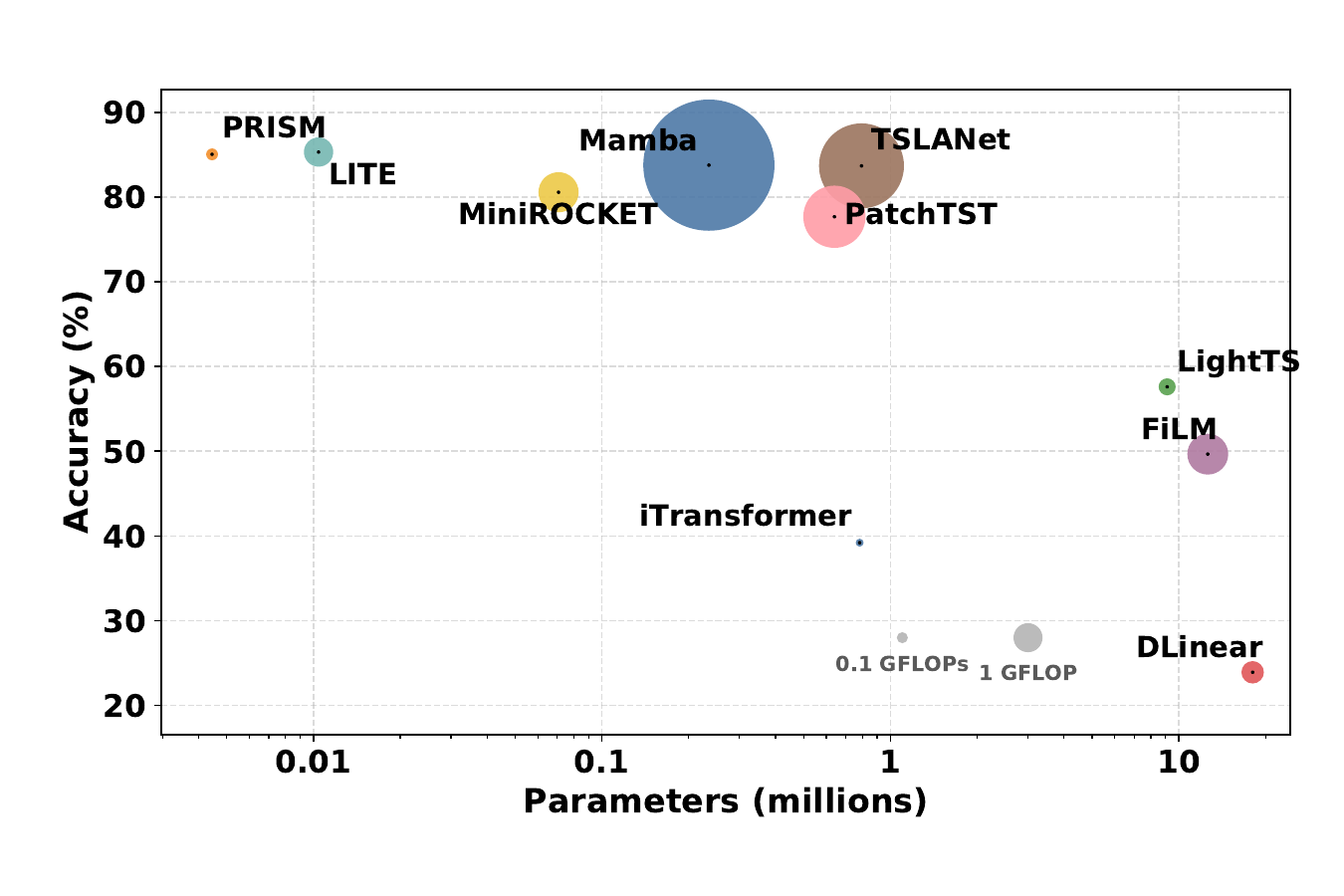}
    \caption{Sleep-EDF — accuracy vs. parameters and complexity}
    \label{fig:accuracy_vs_gflops_sleep_edf}
  \end{subfigure}

  \vspace{0.75em} 

  \begin{subfigure}{0.48\textwidth} 
    \centering
    \includegraphics[width=\linewidth]{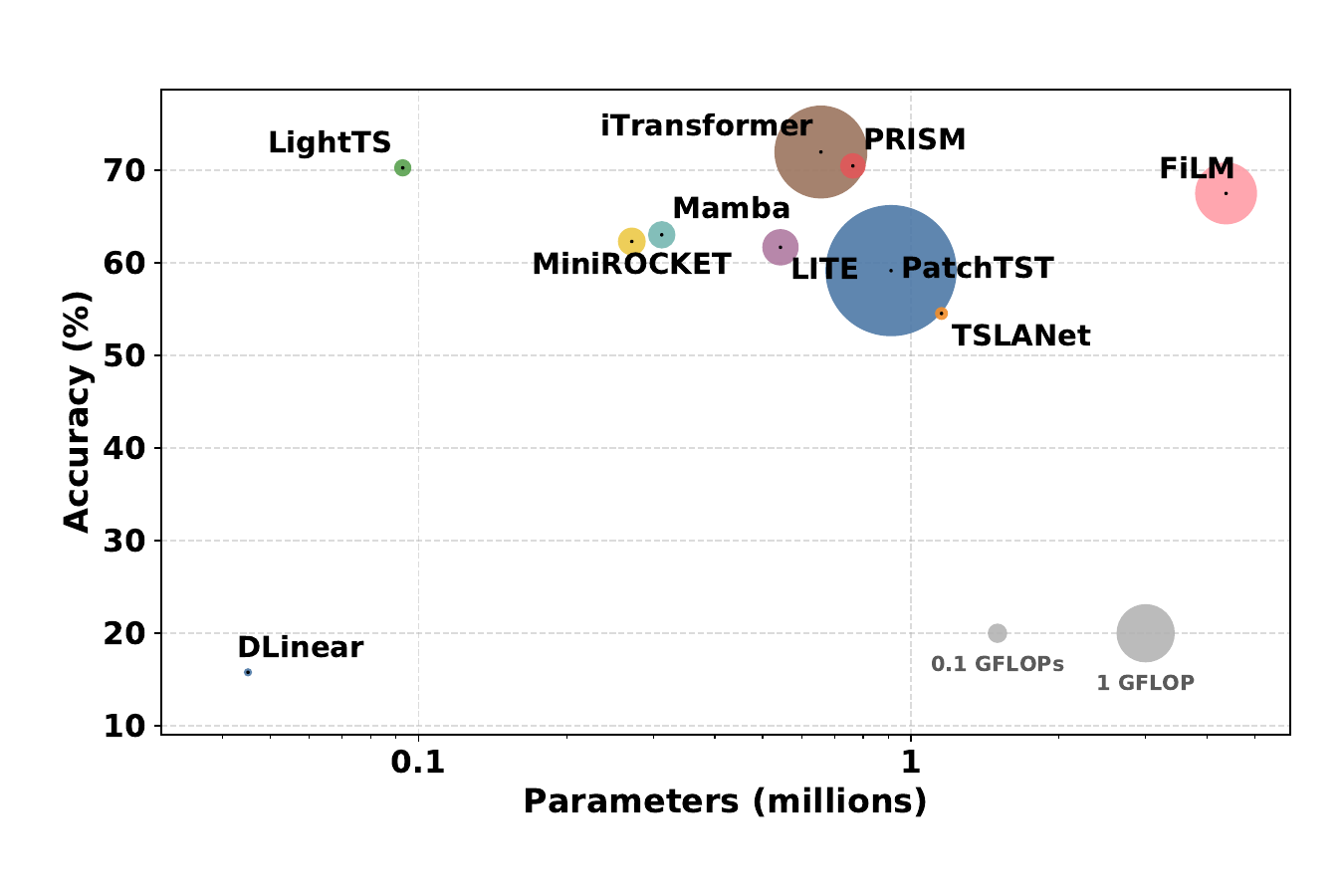}
    \caption{InsectWingbeat — accuracy vs. parameters and  complexity}
    \label{fig:accuracy_vs_gflops_insectwingbeat}
  \end{subfigure}

    \caption{Accuracy vs.\ complexity for PRISM and baselines on (a) UCI-HAR, (b) Sleep-EDF, and (c) InsectWingbeat. The marker size is proportional to the computational cost, measured in GFLOPs, providing a visual indication of each model's complexity.}

  \label{fig:accuracy_vs_gflops}
\end{figure*}

To provide a comprehensive view of model efficiency, we evaluate the computational footprint of PRISM and the baseline models across both archives. As shown in Table \ref{tab:uea-results}, PRISM operates with an average of 0.59G FLOPs and 66.26K parameters on the UEA archive, standing as the most computationally efficient model in our comparison. It requires roughly a third of the compute of LITE (1.57G FLOPs), significantly less than Mamba (3.82G FLOPs), and over 15$\times$ fewer FLOPs than MiniROCKET (9.34G FLOPs). This trend continues on the biomedical and HAR datasets (Table \ref{tab:human-activity-recognition-and-biomedical-signal-results}), where PRISM achieves an average compute cost of only 0.04G FLOPs, substantially lower than LITE (0.27G FLOPs) and orders of magnitude lower than Mamba (5.11G FLOPs).

In Figure \ref{fig:accuracy_vs_gflops} we compare the accuracy, parameter count, and computational cost of the tested models across three representative datasets from the BioHAR and UEA archive.
Computation cost is conveyed through the marker size, which scales proportionally with the FLOPs.

Focusing on UCI-HAR (Figure \ref{fig:accuracy_vs_gflops_ucihar}), PRISM attains 96.37\% mean accuracy while requiring only 0.048 GFLOPs and 35K parameters, placing it among the most computationally efficient models in the comparison. By contrast, achieving comparable accuracy requires 0.131 GFLOPs with LITE, 0.421 GFLOPs with MiniROCKET, and 0.929 GFLOPs with Mamba. Compared to TSLANet, PRISM operates with roughly 8$\times$ fewer FLOPs and substantially fewer parameters, while still achieving an accuracy gain of around 4\%. In addition, against larger transformer-based models such as iTransformer, PRISM demonstrates a 5.5\% improvement in accuracy alongside a 2.5$\times$ reduction in compute. Overall, these representative cases indicate that PRISM achieves competitive or superior accuracy while maintaining a comparatively small computational footprint.

For Sleep-EDF (Figure \ref{fig:accuracy_vs_gflops_sleep_edf}), where models process substantially longer input sequences, PRISM continues to scale efficiently. It reaches 85.02\% mean accuracy using only 0.12 GFLOPs and 4.5K parameters. In comparison, LITE requires 1.02 GFLOPs to process the same sequences, while Mamba demands 21.47 GFLOPs. TSLANet achieves a similar level of accuracy (83.67\%) but requires nearly 9 GFLOPs and 794K parameters, representing nearly two orders of magnitude higher computational cost. DLinear, although lightweight in FLOPs, exhibits a parameter count that is more than 4,000$\times$ larger than PRISM, while also showing a substantial reduction in accuracy (23.92\%).

Finally, to assess behavior when the input dimensionality becomes substantially larger, we include results on InsectWingbeat (Figure \ref{fig:accuracy_vs_gflops_insectwingbeat}), a dataset with 200 input channels. In this setting, PRISM's parameter count and FLOP budget increase relative to the previous datasets, yet they remain limited in absolute terms: PRISM attains 70.48\% mean accuracy at 0.183 GFLOPs with 762.7K parameters. This remains highly efficient compared to these baselines, consuming less compute than Mamba (0.211 GFLOPs) and roughly half that of LITE (0.383 GFLOPs), even though PRISM requires a higher parameter count in this specific high-dimensional setting compared to LITE (543.5K) and Mamba (311.9K). Compared to iTransformer, PRISM trades a small difference in accuracy (71.98\% vs.\ 70.48\%) for considerably lower computational cost (about 14$\times$ fewer FLOPs; 2.57 vs.\ 0.183 GFLOPs), with a similar parameter scale. Relative to TSLANet, PRISM reaches higher accuracy (70.48\% vs.\ 54.54\%) while requiring fewer parameters (762.7K vs.\ 1.15M). Overall, these results suggest that PRISM is able to accommodate high input dimensionality while keeping computational demands and model size comparatively controlled.

\section{Case Study: ISRUC-S3 Dataset}

While PRISM demonstrates strong performance on widely used public benchmarks, we further evaluate its capabilities on the ISRUC-S3 dataset a multimodal sleep stage classification task that presents significant additional challenges. ISRUC-S3 is not only more complex due to the diversity of sensor modalities involved (EEG, EOG, EMG, ECG), but it also reflects practical constraints commonly encountered in clinical scenarios, such as limited subject numbers.

By evaluating PRISM under these demanding conditions, we aim to assess the architecture’s robustness and understand its suitability for real-world scenarios. This experiment is designed to probe two central aspects of PRISM: (1) how well it integrates and processes information from multiple sensor-specific channels, and (2) how its compact computational footprint, in terms of both MFLOPs and parameter count, positions it for deployment in resource-constrained environments such as wearable devices or bedside monitoring systems.

\subsection{Dataset Description}
The ISRUC-S3 dataset \cite{khalighi2016isruc} consists of polysomnographic recordings from 10 healthy subjects collected during sleep. The signals are segmented into 30-second epochs, yielding a total of 8,589 samples. Sleep stages are annotated by two experts according to the AASM standard, resulting in five classes: Wake, N1, N2, N3, and REM. The dataset includes six referenced EEG channels, two EOG channels, one EMG channel, and one ECG channel, all downsampled to 100 Hz. Consistent with previous work \cite{goerttler2025msa}, the ECG channel is excluded from analysis due to its periodic nature. To suppress high-frequency noise, each channel is preprocessed using a fourth-order low-pass Butterworth filter with a 40 Hz cutoff.

\subsection{Experimental Protocol}
The evaluation on ISRUC-S3 follows a 10-fold repeated subject-wise cross-validation, as recommended in prior work. Given the small number of subjects, cross-validation is repeated 10 times with fixed folds across repetitions to ensure robust estimates despite potential overlap between test subjects across folds. This protocol addresses the reduced effective sample size and provides a fair basis for model comparison.

To ensure consistency and facilitate a direct comparison, all experiments and performance assessments are conducted within the same framework and evaluation pipeline as the original MSA-CNN study \cite{goerttler2025msa}. The results reported for comparative methods are therefore taken directly from their study \cite{goerttler2025msa}.

Analogous to the ``small'' configuration explored for MSA-CNN study, we also evaluated a lightweight variant of PRISM on the ISRUC-S3 dataset: convolutional kernel sizes of 7, 15, and 25 in the temporal-convolution blocks, an embedding dimension of 26 in the final convolutional layer, and depth-wise convolutional layers with a kernel size of 8 at intermediate stages.

\subsection{Evaluation Metrics}
Model performance is reported using three complementary metrics:
\begin{itemize}[leftmargin=*, topsep=0pt, itemsep=0pt]
    \item \textbf{Accuracy:} The proportion of correctly classified samples across all sleep stages.
    \item \textbf{Macro F1 Score:} The unweighted mean of F1-scores computed per class, which mitigates class imbalance effects.
    \item \textbf{Cohen's Kappa:} Measures the agreement between predicted and true labels while correcting for chance, providing a robust assessment in the presence of class imbalance.
\end{itemize}
For each metric, results are first averaged over test folds (weighted by the number of samples per fold), then averaged across all repetitions.

\subsection{Baselines}
PRISM is tested against state-of-the-art methods specifically developed for sleep stage classification using raw EEG signals, without relying on hand-crafted features. These include convolutional and recurrent models such as DeepSleepNet \cite{supratak2017deepsleepnet}, which combines CNNs for feature extraction with bidirectional LSTMs for modelling temporal transitions, and EEGNet \cite{lawhern2018eegnet}, which uses depthwise-separable convolutions for efficient temporal and spatial filtering. Other approaches, like AttnSleep \cite{eldele2021attention}, integrate multi-resolution CNNs with attention mechanisms to capture both frequency-specific patterns and long-range dependencies. A distinct class of models, including MSTGCN \cite{jia2021multi}, JK-STGCN \cite{ji2022jumping}, HierCorrPool \cite{wang2023multivariate}, and FC-STGNN \cite{wang2024fully}, leverages graph neural networks to encode spatial and temporal relationships across EEG channels or multimodal inputs. Finally, models like cVAN \cite{yang2024cvan} and MSA-CNN \cite{goerttler2025msa} combine CNN-based spectral encoders with Transformer-like attention blocks to align and contextualise temporal representations at multiple scales.

\subsection{Results and Discussion}

\begin{table}[tb]
\centering
\caption{Classification results on the ISRUC-S3 dataset (multivariate) in terms of mean $\pm$ std (\%). The best performance is highlighted in \textcolor{blue}{\textbf{blue}}, and the second-best result is \textcolor[HTML]{D55E00}{\underline{orange}}.}
\label{tab:isruc-s3-results}
\resizebox{0.5\linewidth}{!}{
\begin{tabular}{l|cccc}
\toprule
Model         & Accuracy       & Macro F1        & Kappa           \\
\midrule
EEGNet~              & $75.9 \pm 1.0$          & $71.9 \pm 1.0$           & $68.2 \pm 1.3$           \\
GraphSleepNet~         & $69.8 \pm 1.7$          & $66.7 \pm 2.1$           & $60.3 \pm 2.1$           \\
MSTGCN          & $77.2 \pm 0.6$          & $73.5 \pm 0.7$           & $69.7 \pm 0.7$           \\
JK-STGCN         & $75.9 \pm 0.6$          & $72.0 \pm 0.4$           & $68.1 \pm 0.7$           \\
HierCorrPool    & $72.1 \pm 1.3$          & $67.8 \pm 1.6$           & $63.1 \pm 1.7$           \\
FC-STGNN        & $68.6 \pm 0.4$          & $64.6 \pm 1.0$           & $58.5 \pm 0.8$           \\
cVAN           & $72.5 \pm 2.0$          & $67.7 \pm 2.5$           & $63.7 \pm 2.5$           \\
MSA-CNN (small)           & \textcolor{blue}{\textbf{$79.8 \pm 0.9$}}   & \textcolor{blue}{\textbf{$76.8 \pm 1.1$}}    & \textcolor{blue}{\textbf{$73.2 \pm 1.1$}}    \\
\midrule
 \textbf{PRISM (small)}         &  \textcolor[HTML]{D55E00}{\underline{$78.1 \pm 0.5$}}  &  \textcolor[HTML]{D55E00}{\underline{$74.8 \pm 0.6$}} &  \textcolor[HTML]{D55E00}{\underline{$71.1 \pm 0.6$}} \\
\bottomrule
\end{tabular}
}
\end{table}

\begin{table}[tb]
  \centering
  \caption{Comparison of model parameters and computational cost (MFLOPs) on the ISRUC-S3 dataset. The best result (lowest) is highlighted in \textcolor{blue}{\textbf{blue}}, and the second-best result is \textcolor[HTML]{D55E00}{\underline{orange}}.}
  \label{tab:isruc_multivariate}
  \resizebox{0.5\linewidth}{!}{
  \begin{tabular}{lrr}
  
    \toprule
    Model & \# Parameters & MFLOPs \\
    \midrule
    EEGNet          & 12{,}181        & 32.8                \\
    MSTGCN          & 476{,}292       & 204.0              \\
    JK-STGCN        & 458{,}085       & 208.6              \\
    HierCorrPool    & 13{,}290{,}000  & 499.0              \\
    FC-STGNN        & 3{,}200{,}000   & 475.8              \\
    cVAN            & 6{,}834{,}571   & 483.4              \\
    MSA-CNN (small) & \textcolor[HTML]{D55E00}{\underline{10{,}583}}  & \textcolor[HTML]{D55E00}{\underline{\textcolor[HTML]{D55E00}{19.8}}}  \\
    \midrule
    \textbf{PRISM (small)} & \textbf{\textcolor{blue}{3{,}817}} & \textbf{\textcolor{blue}{8.3}} \\
    \bottomrule
  \end{tabular}
  }

\end{table}

Table \ref{tab:isruc-s3-results} reports the classification performance of PRISM alongside a range of established baselines on the ISRUC-S3 multivariate dataset, while Table \ref{tab:isruc_multivariate} summarises the corresponding parameter counts and computational complexity. Together, these tables form the basis for evaluating both predictive performance and computational efficiency.

PRISM achieves performance that is competitive with the strongest baselines. Although MSA-CNN (small) attains the highest accuracy ($79.8\%\pm0.9$), macro F1 ($76.8\%\pm1.1$), and Cohen's $\kappa$ ($73.2\%\pm1.1$), PRISM remains within approximately two percentage points across all metrics, achieving $78.1\%\pm0.5$ accuracy, $74.8\%\pm0.6$ macro F1, and $71.1\%\pm0.6$ $\kappa$. Moreover, PRISM surpasses several substantially larger and more complex architectures, including EEGNet ($75.9\%$ accuracy), GraphSleepNet ($69.8\%$), JK-STGCN ($75.9\%$), HierCorrPool ($72.1\%$), FC-STGNN ($68.6\%$), and cVAN ($72.5\%$), across all reported performance measures.

Table \ref{tab:isruc_multivariate} illustrates the distinct computational advantages of PRISM. With only 3,817 parameters and 8.3 MFLOPs, it is the most lightweight model evaluated; even the compact MSA-CNN (small) requires nearly three times as many parameters and more than twice the computational cost. The contrast is particularly striking against graph-based architectures: PRISM outperforms HierCorrPool and cVAN by roughly six percentage points in accuracy while using less than 0.03\% of the parameters (3,817 vs.\ 13.29M for HierCorrPool) and approximately 1.7\% of the computational cost. This demonstrates that PRISM achieves a favourable balance between accuracy and efficiency that is not realised by the more resource-intensive baselines.

\section{Model Analysis}
In this section, we perform a detailed analysis to validate the architectural choices underpinning PRISM. We begin by examining the scalability of the multi-resolution design, quantifying the trade-offs between classification accuracy and computational complexity as the number of temporal scales and kernels varies. Subsequently, we investigate the specific impact of the symmetry constraint on the learned representations. By comparing symmetric and asymmetric filters, we analyse their spectral characteristics, specifically focus, attenuation, and diversity to understand how this structural prior influences the model's ability to capture discriminative temporal features.

\subsection{Ablation On The Number of Resolutions and Kernels Per Resolution}
\begin{figure}[tb]
    \centering
    \includegraphics[width=0.8\linewidth]{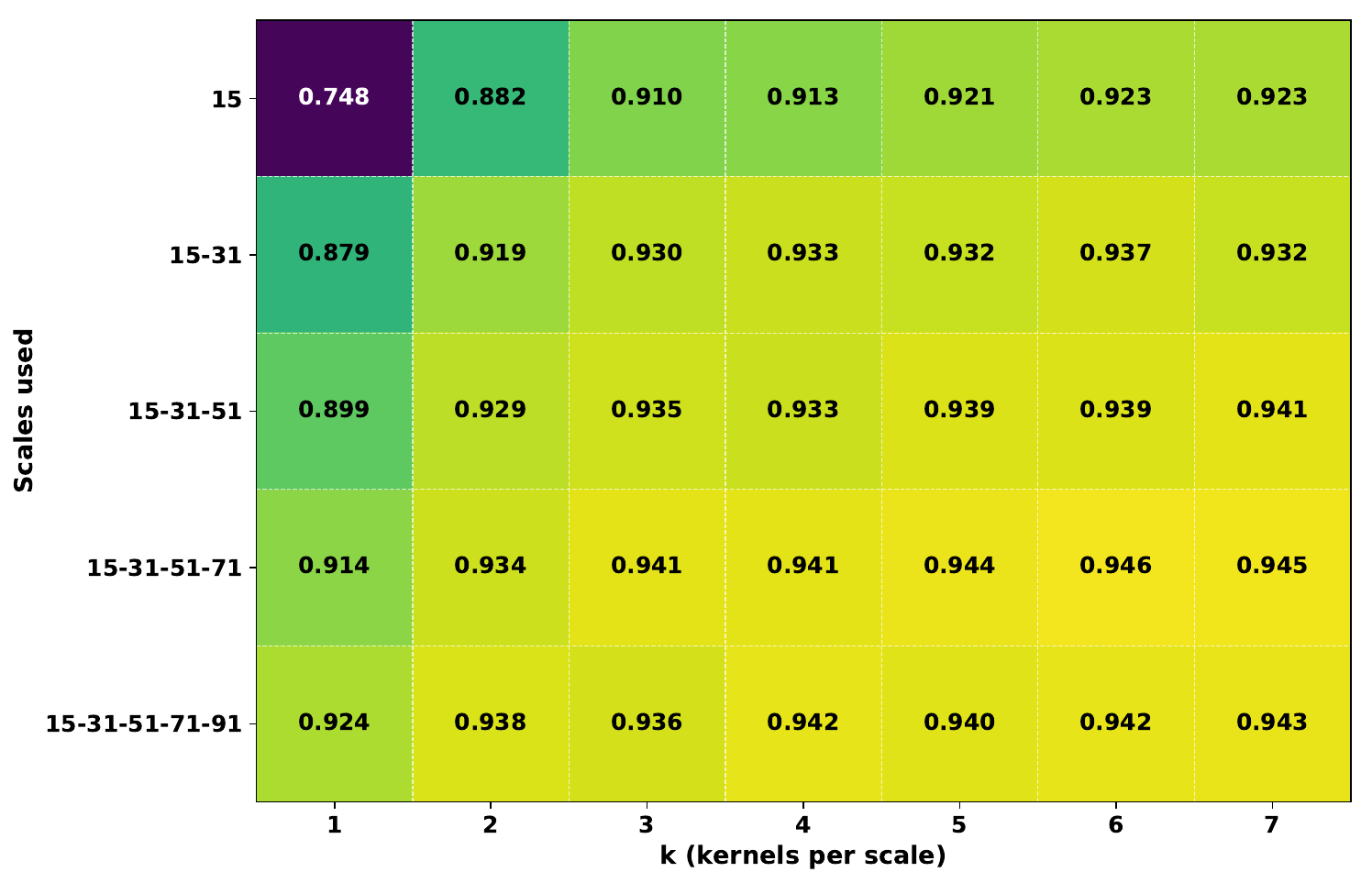}
    \caption{Heatmap of the mean classification accuracy, where each cell represents the average performance across all HAR and BIO datasets, as a function of the scale set (rows) and the number of kernels per scale (k) (columns). Each cell is annotated with the corresponding averaged accuracy.}
    \label{fig:kernel_ablation_heatmap_biohar}
\end{figure}

We evaluated how the architecture scales with different combinations of temporal resolutions and kernels per resolution. The goal was to determine whether performance gains arise primarily from using additional resolutions, which extend the receptive field, or from increasing the number of kernels, which expands the representational capacity at each scale. All other hyperparameters were kept fixed to ensure a fair comparison across configurations.

Figure \ref{fig:kernel_ablation_heatmap_biohar} reports the mean classification accuracy across all HAR and BIO datasets as a function of the number of temporal scales (rows) and the number of kernels per scale \(k\) (columns). The most pronounced improvement is observed when introducing a second temporal scale while keeping \(k=1\), where accuracy increases from 74.8\% to 87.9\% (+13.1 p.p.). This highlights the critical role of multi-resolution processing, as even a single additional receptive field substantially enhances performance. 

Beyond this point, the effect of adding further scales becomes progressively smaller. At \(k=5\), accuracy increases only modestly from 92.3\% to 94.6\% when moving from one to four scales, before slightly decreasing to 94.2\% with the inclusion of a fifth scale. This suggests that, after a certain receptive-field diversity is reached, additional scales contribute redundant information rather than complementary features.

Increasing the number of kernels per scale yields consistent but diminishing gains. For a single scale, performance rises from 74.8\% to 92.3\% (+17.5 p.p.) as \(k\) increases from 1 to 5. For two scales, the improvement is reduced to +5.8 p.p.\ (87.9\% to 93.7\%), and continues to decline for higher numbers of scales: +4.0  p.p.\ for three scales, +3.2  p.p.\ for four, and only +1.8 p.p.\ for five. These results indicate that once each resolution is sufficiently represented, additional kernels provide limited benefit and may even introduce redundancy.

\begin{figure}[tb]
    \centering
    \includegraphics[width=0.8\linewidth]{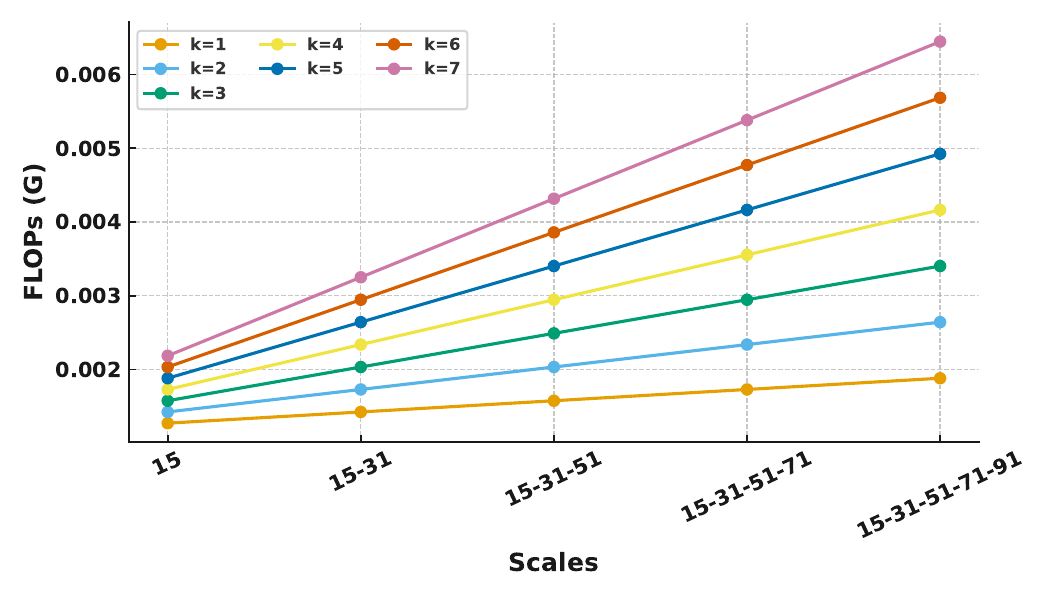}
    \caption{Computational complexity (GFLOPs) as a function of the scale set and the number of kernels per scale $k$. Each curve corresponds to a fixed $k$ and illustrates how the overall cost increases when adding new temporal resolutions}
    \label{fig:compute_vs_scales_fixed_k}
\end{figure}

Figure \ref{fig:compute_vs_scales_fixed_k} shows that computational cost increases linearly with both the number of temporal resolutions and the number of kernels per scale. Moving from one to five scales triples to quadruples the FLOPs, since each added scale introduces an additional set of \(k\) convolutions. When combined with the results in Figure \ref{fig:kernel_ablation_heatmap_biohar}, this reveals a clear efficiency saturation: for instance, increasing from three to five scales raises computation by more than 80\% while improving accuracy by only 0.3 percentage points (from 93.9\% to 94.2\%), confirming that additional resolutions yield diminishing returns beyond moderate depth.

Overall, the findings demonstrate that both temporal resolution diversity and kernel multiplicity contribute meaningfully to performance improvements. The gain from additional scales is more substantial, but increasing the number of kernels per scale also strengthens representational capacity up to a moderate level. The optimal trade-off between accuracy and efficiency is achieved with three to four scales and approximately five kernels per scale, beyond which performance stabilises and further architectural growth yields diminishing returns.

\subsection{Sensitivity Analysis for Selecting $n_k$ and $n_f$}\label{subsec:sensitivity_nk_nf}
To provide practical guidance for choosing the number of temporal scales ($n_k$) and the number of filters per scale ($n_f$), we quantify how test accuracy varies across the ablation grid. For each dataset, we compute the Spearman rank correlation between the hyperparameter value and test accuracy, i.e., $\rho(n_k,\mathrm{Acc})$ and $\rho(n_f,\mathrm{Acc})$. These correlations act as simple sensitivity scores: a large positive $\rho$ indicates that increasing $n_k$ or $n_f$ tends to yield consistent improvements, whereas small or negative values suggest diminishing returns and/or overfitting.

\begin{figure}[tb]
    \centering
    \begin{subfigure}[t]{0.49\columnwidth}
        \centering
        {\color{blue}\includegraphics[width=\linewidth]{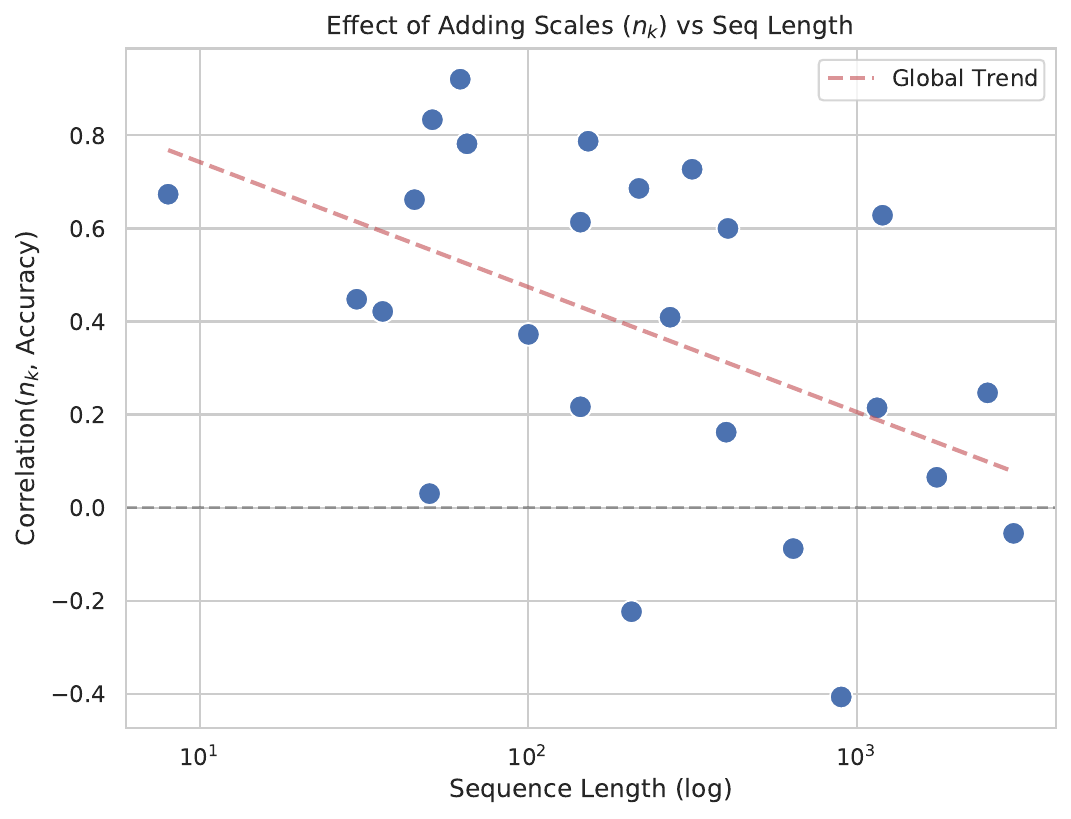}}
        \caption{Scales vs.\ length}
        \label{fig:trend_scales_vs_length_uea}
    \end{subfigure}\hfill
    \begin{subfigure}[t]{0.49\columnwidth}
        \centering
        {\color{blue}\includegraphics[width=\linewidth]{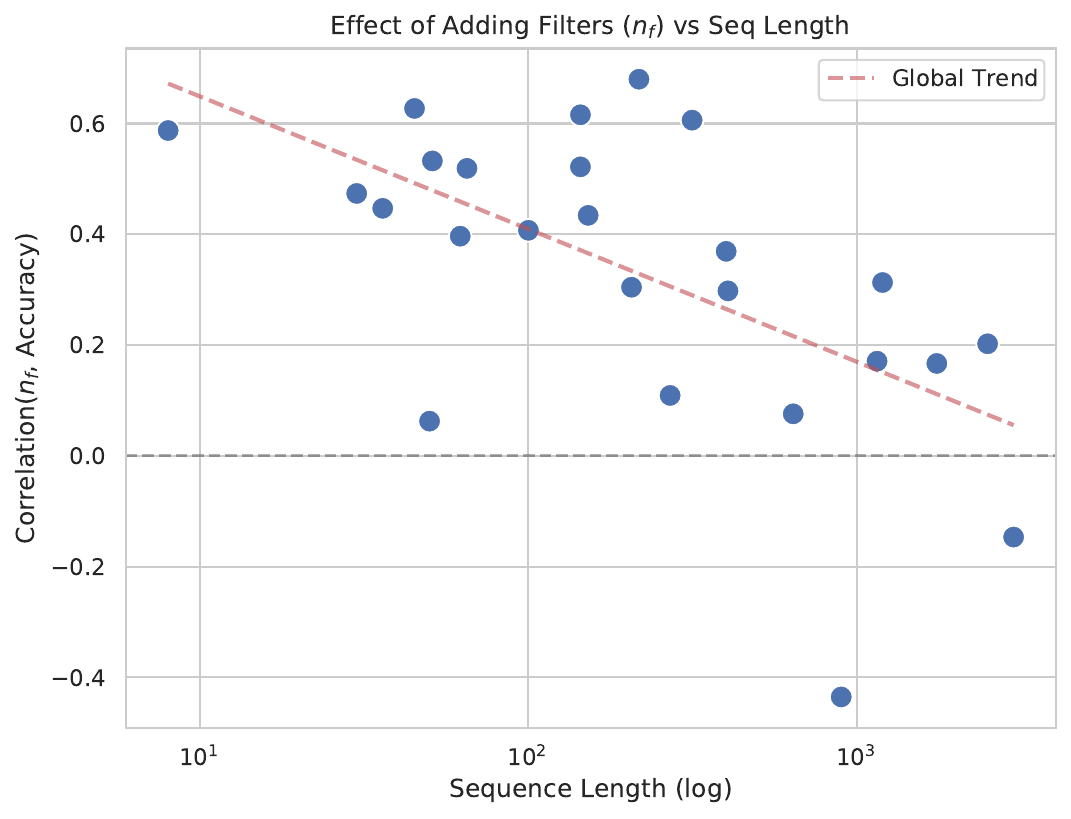}}
        \caption{Filters vs.\ length}
        \label{fig:trend_filters_vs_length_uea}
    \end{subfigure}
    \caption{Sensitivity of test accuracy to the number of scales ($n_k$) and filters per scale ($n_f$) as a function of dataset sequence length $L$ (log $x$-axis). Trend lines are fitted in $\log_{10}(L)$.}
    \label{fig:sensitivity_trends}
\end{figure}

Figure~\ref{fig:sensitivity_trends} relates these per-dataset sensitivities to sequence length $L$. Specifically, it plots $\rho$ against $L$ using a logarithmic $x$-axis, with a global trend line in $\log_{10}(L)$ to summarise the overall tendency. The resulting meta-analysis reveals an overall diminishing-return effect with increasing sequence length: scales vs.\ length $\rho=-0.463$ ($p=0.0198$) and filters vs.\ length $\rho=-0.593$ ($p=0.0018$). Furthermore, observing the trend lines at short sequence lengths (e.g., $L < 10^2$) reveals that the baseline sensitivity to scales ($\rho \approx 0.8$) is notably higher than the baseline sensitivity to filters ($\rho \approx 0.65$).
Overall, these results support a simple selection strategy. For shorter sequences, adding temporal scales (increasing $n_k$) tends to produce larger gains, and we therefore recommend prioritising scale diversity before increasing the number of filters per scale. For longer sequences and/or higher-dimensional datasets, the incremental benefit of enlarging $n_k$ or $n_f$ decreases, so compact configurations are often sufficient; in these settings, $n_k$ can be kept modest and $n_f$ should be increased only if underfitting is observed.

\begin{figure*}[t]
    \centering
    \begin{subfigure}[t]{0.32\textwidth}
        \centering
        \includegraphics[width=\linewidth]{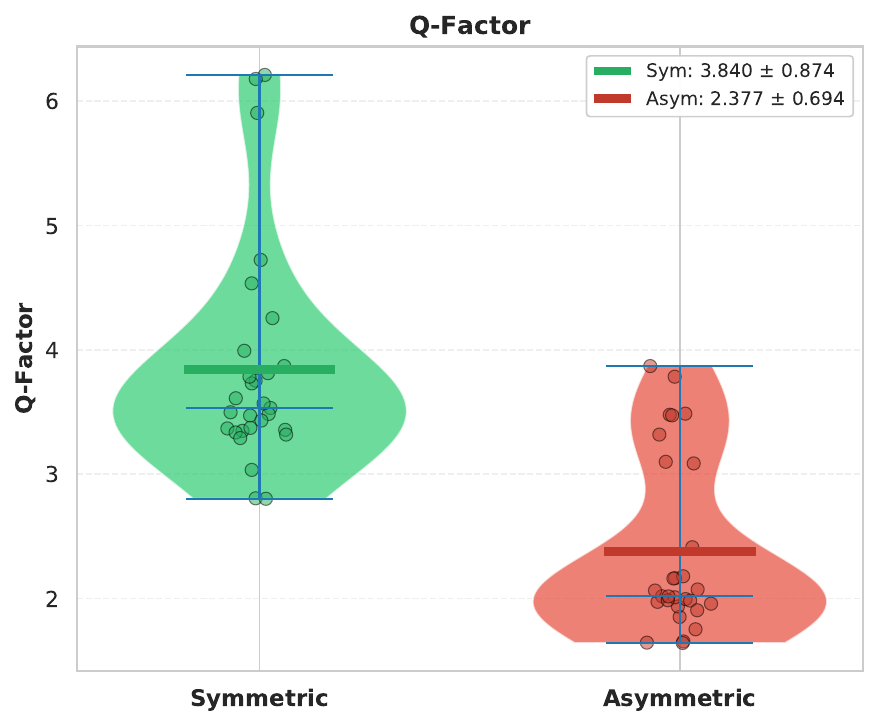}
        \caption{Q-factor comparison}
        \label{fig:spectral_q_factor}
    \end{subfigure}
    \hfill
    \begin{subfigure}[t]{0.32\textwidth}
        \centering
        \includegraphics[width=\linewidth]{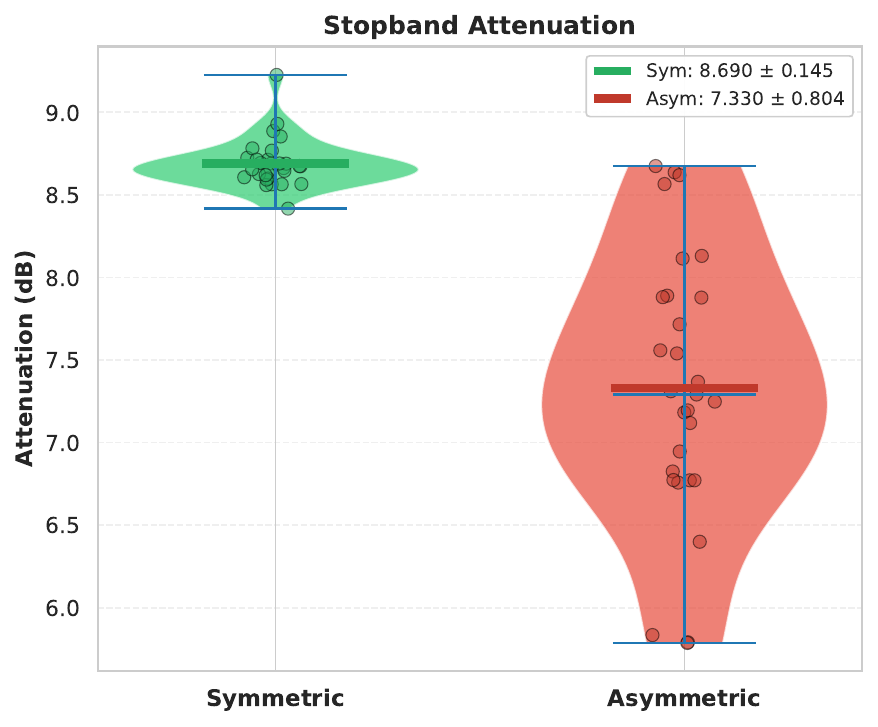}
        \caption{Stopband attenuation}
        \label{fig:spectral_stopband_attenuation}
    \end{subfigure}
    \hfill
    \begin{subfigure}[t]{0.32\textwidth}
        \centering
        \includegraphics[width=\linewidth]{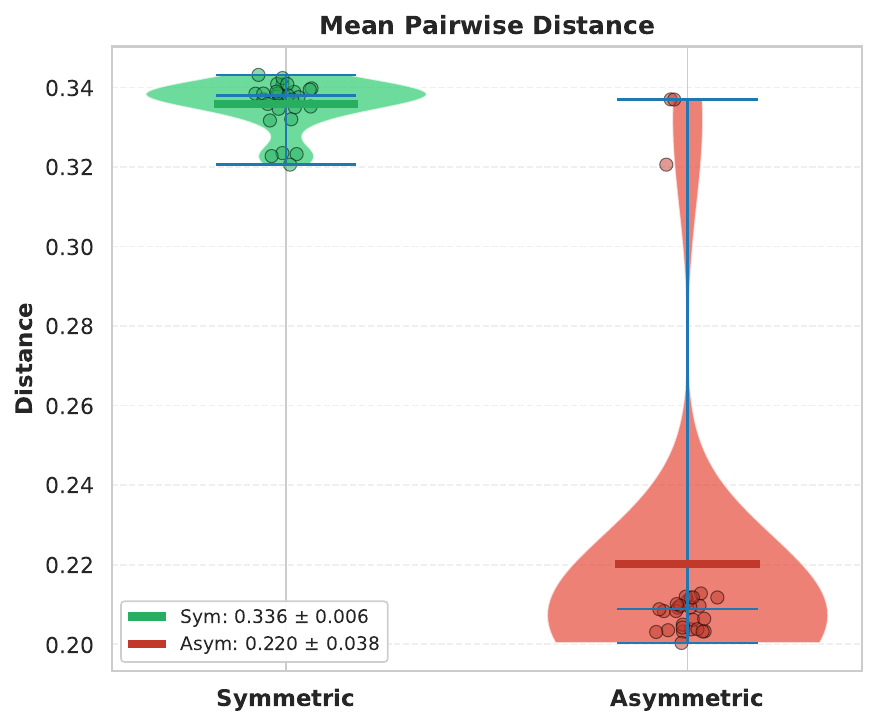}
        \caption{Pairwise spectral distance}
        \label{fig:spectral_pairwise_distance}
    \end{subfigure}

    \caption{
    Comparison of spectral metrics between symmetric and asymmetric PRISM filters across 29 UEA datasets.  
    The three panels report Q-factor, stopband attenuation, and pairwise spectral distance respectively.
    }
    \label{fig:spectral_metrics_violin}
\end{figure*}

\subsection{Effect of Symmetric Filtering on Filter Selectivity, Diversity, and Perfomance}\label{subsec:symmetric_filter_analysis}

To study how the symmetry constraint shapes the spectral structure of learned filters and impacts classification performance, we compared convolutional filters trained with and without symmetric weight sharing. Both variants used identical architectures and training settings so that any difference arises from symmetry alone.

\subsubsection{Spectral Metrics}

\paragraph{Q-Factor (Filter Selectivity)} measures how concentrated a filter is around its dominant frequency. It is defined as the ratio between the centre frequency \(f_c\) and the effective three decibel bandwidth, such that
\[
Q = \frac{f_c}{\text{BW}}.
\]
Larger values indicate sharper and more selective spectral responses.

\paragraph{Stopband Attenuation} quantifies how strongly a filter suppresses frequencies outside its passband. It is computed as the mean magnitude of the spectrum in regions that lie at least twenty decibels below the peak. Higher attenuation corresponds to cleaner, less noisy spectral envelopes.

\paragraph{Frequency Diversity} measures how different the filters are from one another in the spectral domain. It is computed as the mean pairwise cosine distance between normalised magnitude spectra. Larger distances imply that filters spread more uniformly across the frequency axis rather than collapsing onto similar frequency responses.

\subsubsection{Analysis Protocol and Results}

For each of the 29 UEA datasets, we extracted convolutional filters from both symmetric and asymmetric models. After applying the fast Fourier transform and computing magnitude spectra in decibels, we evaluated Q-factor, stopband attenuation, and pairwise spectral distance for every filter. These values were aggregated per dataset, and paired comparisons across datasets were assessed with the Wilcoxon signed-rank test. Effect sizes were computed with Cliff's delta.

\begin{description}[leftmargin=*,topsep=2pt,itemsep=2pt]

\item[\textbf{(i) Filter Selectivity}]  
The comparison in Fig.\ \ref{fig:spectral_q_factor} shows that symmetric filters exhibit substantially higher Q-factors than asymmetric ones. Across the archive, the symmetric variant has a mean Q-factor of \(3.8396 \pm 0.88\) compared with \(2.3774 \pm 0.75\) for the asymmetric model (\(p = 5.903\times10^{-6}\), \(\delta = 0.790\)). Symmetry therefore sharpens the passband and reduces spectral spread.

\item[\textbf{(ii) Stopband Attenuation}]  
Figure \ref{fig:spectral_stopband_attenuation} illustrates that symmetric filters achieve stronger attenuation outside the main lobe. Their mean stopband level is \(-8.6903 \pm 0.36\) decibels, compared with \(-7.3303 \pm 1.20\) decibels for the asymmetric filters (\(p = 4.149\times10^{-6}\), \(\delta = 0.906\)). The symmetry constraint encourages smoother, more predictable spectral shapes similar to linear phase FIR filters, improving noise suppression and temporal stability.

\begin{figure}[tb]
    \centering
    \includegraphics[width=0.8\linewidth]{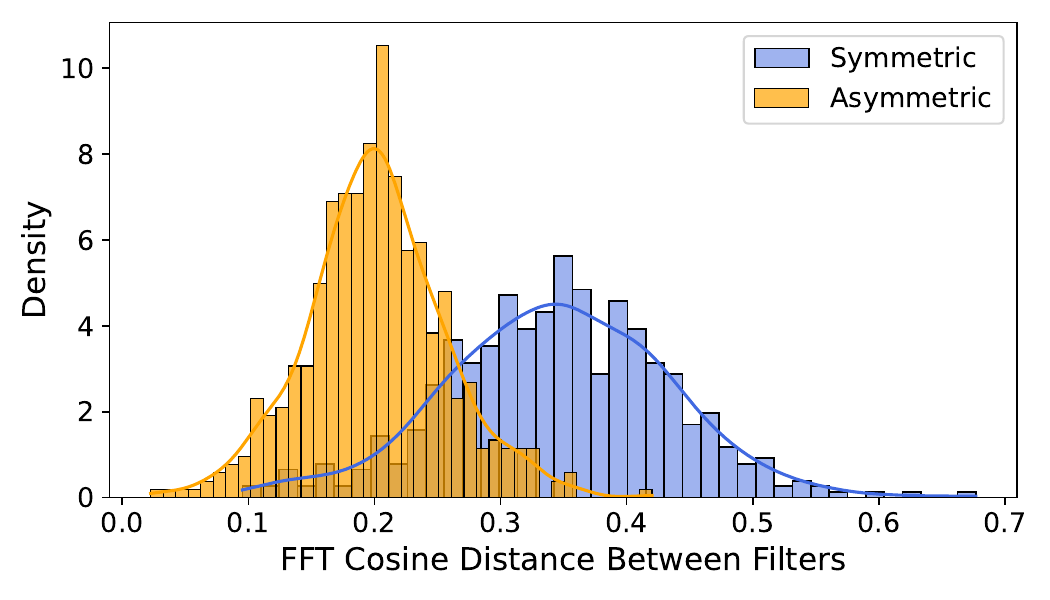}
    \caption{
    Distribution of pairwise cosine distances between FFT magnitude spectra of learned filters for symmetric and asymmetric convolutions on the WISDM dataset.
    }
    \label{fig:wisdm_filter_pairwise_fft_cosine_distribution}
\end{figure}

\begin{figure}[tb]
    \centering
    \includegraphics[width=\linewidth]{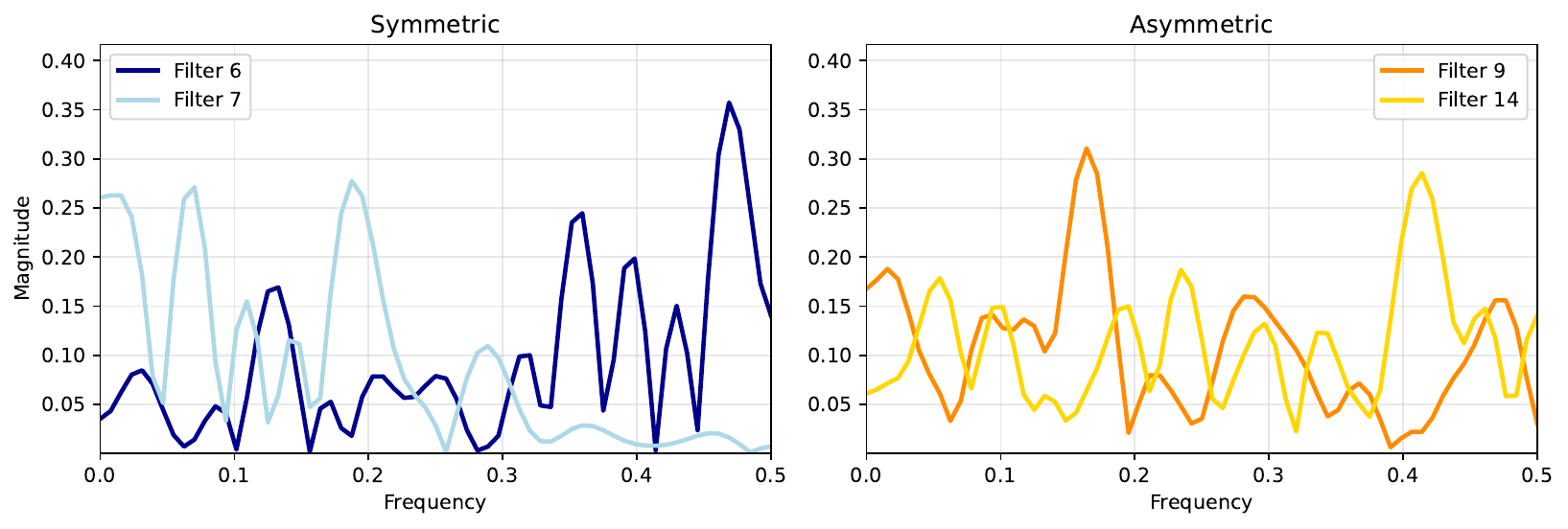}
    \caption{
    Frequency responses of the most spectrally dissimilar filter pairs for kernel size 31 in the symmetric (left) and asymmetric (right) models.
    }
    \label{fig:wisdm_filter_most_different_fft}
\end{figure}

\item[\textbf{(iii) Frequency Diversity}]

Figure \ref{fig:spectral_pairwise_distance} shows that symmetric filters cover the frequency axis more evenly and with less redundancy. The average pairwise cosine distance is \(0.3358 \pm 0.006\) for symmetric filters and \(0.2202 \pm 0.038\) for the asymmetric model (\(p = 4.149\times10^{-6}\), \(\delta = 0.946\)).

On WISDM dataset, the distribution of distances in Figure~\ref{fig:wisdm_filter_pairwise_fft_cosine_distribution} confirms this behaviour. 
The symmetric model reaches a mean of $0.345$ and median of $0.347$, whereas the asymmetric model yields $0.201$ and $0.202$ respectively.

To further illustrate this diversity, we selected the pair of filters with the largest cosine distance between their normalised FFT magnitude spectra for kernel size 31. 
Figure~\ref{fig:wisdm_filter_most_different_fft} visualises these pairs. 
For the symmetric model, filters 6 and 7 were identified as the most different (cosine similarity $0.323$), exhibiting distinct, well-separated frequency profiles, particularly showing a clear separation between high- and low-frequency selectivity. 
In contrast, for the asymmetric model, filters 9 and 14 formed the most dissimilar pair yet still retained a similarity of $0.643$, presenting considerable overlap across the frequency spectrum. To further analyse the differences in the learned filters across multiple channels and show that each channel learns a different representation, we have included a detailed analysis in \ref{appendix:freq_adaptation}.
\begin{figure}[t]
\centering
\includegraphics[width=\linewidth]{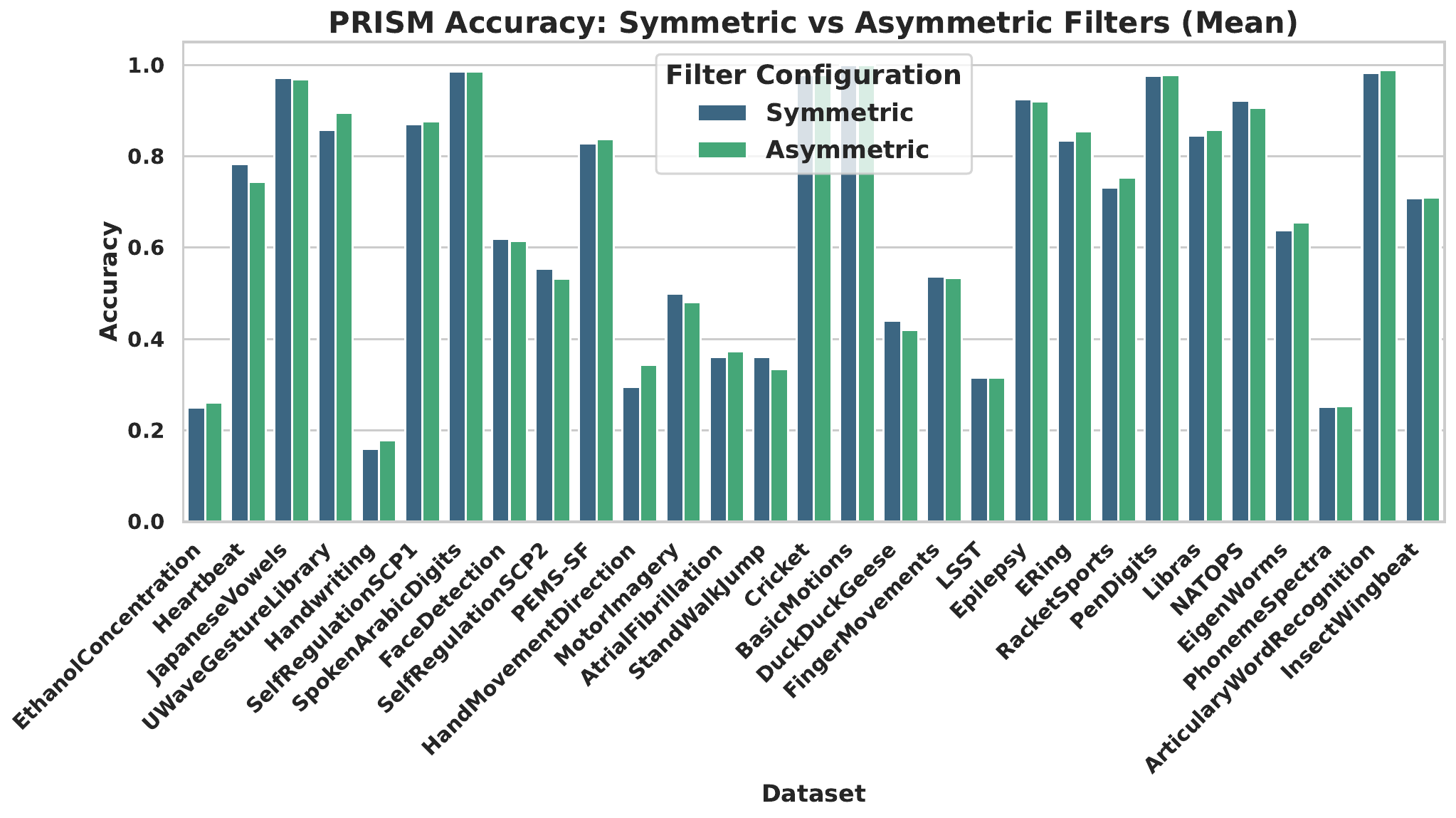}
\caption{PRISM accuracy comparison on UEA datasets using symmetric vs. asymmetric (unconstrained) filters (mean  over 5 repeated runs).}
\label{fig:sym_vs_asym_acc}
\end{figure}
\item[\textbf{(iv) Classification Accuracy}]
Finally, we assess whether the symmetry constraint compromises downstream performance. The results, averaged over 5 runs, shown in Figure~\ref{fig:sym_vs_asym_acc} demonstrate no consistent loss of accuracy for symmetric filters across the UEA datasets. This confirms that the model retains the benefits of parameter efficiency and spectral diversity without sacrificing the discriminative capacity required for classification. For instance, while we observe a minor advantage for symmetric filters on highly periodic data such as the Heartbeat dataset, and a slight disadvantage on datasets such as HandMovementDirection, these variations are highly dataset-specific. Across our experiments, we do not observe a consistent categorical difference between periodic and non-stationary signals.

\end{description}

In summary, symmetry produces filters that are sharper, cleaner, and more diverse in the spectral domain. Higher Q-factors, improved stopband suppression, and broader inter filter diversity collectively show that the symmetry constraint guides learning toward frequency decompositions that are both efficient and interpretable.

\section{Conclusions}

This work introduced PRISM, a fully convolutional classifier for multivariate time series that applies a simple structural prior through symmetric multi-resolution filters. By enforcing symmetry and distributing filters across multiple temporal scales, PRISM constrains the parameterisation of early convolutions while retaining expressive receptive fields. This design achieves substantial parameter and compute reductions and offers a lightweight model suitable for a wide range of applications.

Across diverse benchmarks spanning human activity recognition, biomedical signals, and general multivariate datasets, PRISM matched or surpassed the performance of considerably larger convolutional and Transformer based models. The results show that using multiple filters with different kernel sizes enables a simple linear projection to mix information across resolutions and extract meaningful multi-scale features, and that adopting a channel-independent hypothesis does not significantly limit performance in practice. 

The accompanying ablations clarified how the model benefits from its multi-resolution structure and from the use of symmetric filters. In our experiments symmetric filtering increased selectivity, strengthened attenuation of non-informative frequency content, and encouraged greater spectral diversity across filters. These effects demonstrate that the constraint does more than compress parameters; it influences how filters distribute their spectral responses and how temporal structure is captured across scales. Varying the number and size of resolutions further showed that most gains arise from adding a modest range of temporal scales, while deeper sets of resolutions yield diminishing returns.

While PRISM achieves high performance and efficiency by processing each sensor channel independently, this design inherently omits explicit cross-channel interactions. Although this independence contributes to the model’s lightweight structure and scalability, certain applications may benefit from selectively integrating cross-channel coupling to capture inter-modal dependencies. In this context, methods that combine channel-independent (CI) and channel-dependent (CD) processing offer a practical compromise, consistent with our CI vs.\ CD analysis: the frontend can remain CI to preserve efficiency and regularisation, while later stages introduce lightweight CD mixing to model inter-sensor structure only when it is informative. Thus, exploring mechanisms for efficient cross-channel interaction, without sacrificing PRISM’s computational and parameter efficiency, represents a promising direction for future research. Another promising avenue for improvement lies in the area of self-supervised pre-training. Recent advances in time-series modelling have shown the value of leveraging large corpora of unlabeled data through objectives such as masked-segment reconstruction or contrastive learning \cite{eldele2021tstcc,yue2022ts2vec,meng2023mhccl}. Adapting such self-supervised objectives for PRISM could further improve initialisation, accuracy, and sample efficiency, particularly in data-scarce scenarios.

Altogether, PRISM demonstrates that combining structured convolutional priors with a multi-resolution design can yield an efficient and accurate fully convolutional classifier for time series data, and the results further indicate that classical signal-processing constraints offer a compelling path towards optimising how deep learning models learn from temporal data.

\subsection*{Developmental tendencies and challenges}

\paragraph{Developmental tendencies} Current progress in time-series representation learning suggests a continued shift toward architectures that are simultaneously (i) multi-scale, (ii) frequency-aware, and (iii) explicitly designed to reduce computation and memory in the era of large Transformer models (both during training and at inference time). In this direction, structured frontends (e.g., filterbanks, linear-phase/symmetric constraints, and wavelet/Fourier-inspired modules) are increasingly used to inject inductive biases that improve sample efficiency and interpretability. A second tendency is the rise of hybrid CNN--Transformer designs that use Transformers to capture global patterns and relationships among variables, while using CNN-based components to model local features. Rather than simply stacking a convolutional frontend before attention, many recent architectures emphasise deep feature aggregation, where local and global features interact within intermediate layers of the network.

\paragraph{Challenges} Despite these advances, several open challenges remain. While curated benchmarks are useful for rapid comparison, they can fail to differentiate increasingly complex architectures (e.g., by saturating in accuracy, or by being too small/noisy to expose meaningful differences). Progress therefore also requires evaluation on non-standard datasets with sufficient samples for training and testing, and protocols that better reflect real deployment conditions. Second, fully independent per-channel processing (as in PRISM’s design) can under-utilise cross-channel structure in settings where inter-sensor coupling is crucial; yet introducing channel mixing to compensate can create spurious correlations and shortcut cues that do not reflect genuine temporal structure. Third, multi-resolution systems introduce design choices (kernel sizes, number of resolutions, pooling strategies) that can be dataset-dependent; principled selection rules and better reporting of compute/latency are needed to ensure reproducibility and fair comparison.

\section*{Acknowledgments}
The work of Federico Zucchi has received funding from the Horizon Europe research and innovation programme, under grant agreement No 101095436 and Horizon European innovation council under grant agreement No. 190129251.

\FloatBarrier

\appendix

\section{Supplementary Analysis of Symmetric Multi-Resolution Filters}
\label{appendix:main_supplementary}

This appendix provides extended theoretical formulations and qualitative case studies to supplement the architectural description of PRISM. 

\subsection{Mathematical Formulation of Symmetric Convolutional Filters}
\label{appendix:sym_filter_theory}

In Section~\ref{subsec:mksce} of the main text, we introduce the operational symmetry constraint applied.
The main text establishes that this symmetry induces linear-phase behavior and effectively halves the learnable parameters, the formal mathematical proof linking this mechanism to classical signal processing principles is provided below. 

Based on our methodology, the symmetric kernels in PRISM correspond to Type I linear-phase FIR filters (characterised by a real-valued, symmetric impulse response and an odd length). This can be formalized as follow:

Let $K = 2m + 1$ be an odd kernel length, and let the centre index be $m$. The symmetry constraint is defined as:
\begin{align*}
w[n] = w[K-1-n], \quad n=0,1,\dots,K-1.
\end{align*}

For a real and symmetric impulse response, the discrete-time frequency response factors into:
\begin{align*}
H(e^{j\omega}) &= \sum_{n=0}^{K-1} w[n] e^{-j\omega n} \\
&= e^{-j\omega m}\Big( w[m] + 2\sum_{k=1}^{m} w[m+k]\cos(\omega k) \Big),
\end{align*}
where $e^{-j\omega m}$ represents the linear-phase term (a constant group delay of $m$ samples), and the bracketed expression is a real-valued amplitude term (a linear combination of cosine basis functions).

\subsection{Channel-Specific Frequency Representations and Multi-Resolution Adaptation}
\label{appendix:freq_adaptation}

While the primary text focuses on the numerical evaluation of PRISM using standard datasets, detailing every filter's effect requires extensive visual analysis. To complement the quantitative results, this qualitative case study illustrates how the multi-resolution filters adapt to specific channels.

\begin{figure}[htbp]
    \centering
    \begin{subfigure}[b]{0.49\linewidth}
        \centering
        \includegraphics[width=\linewidth]{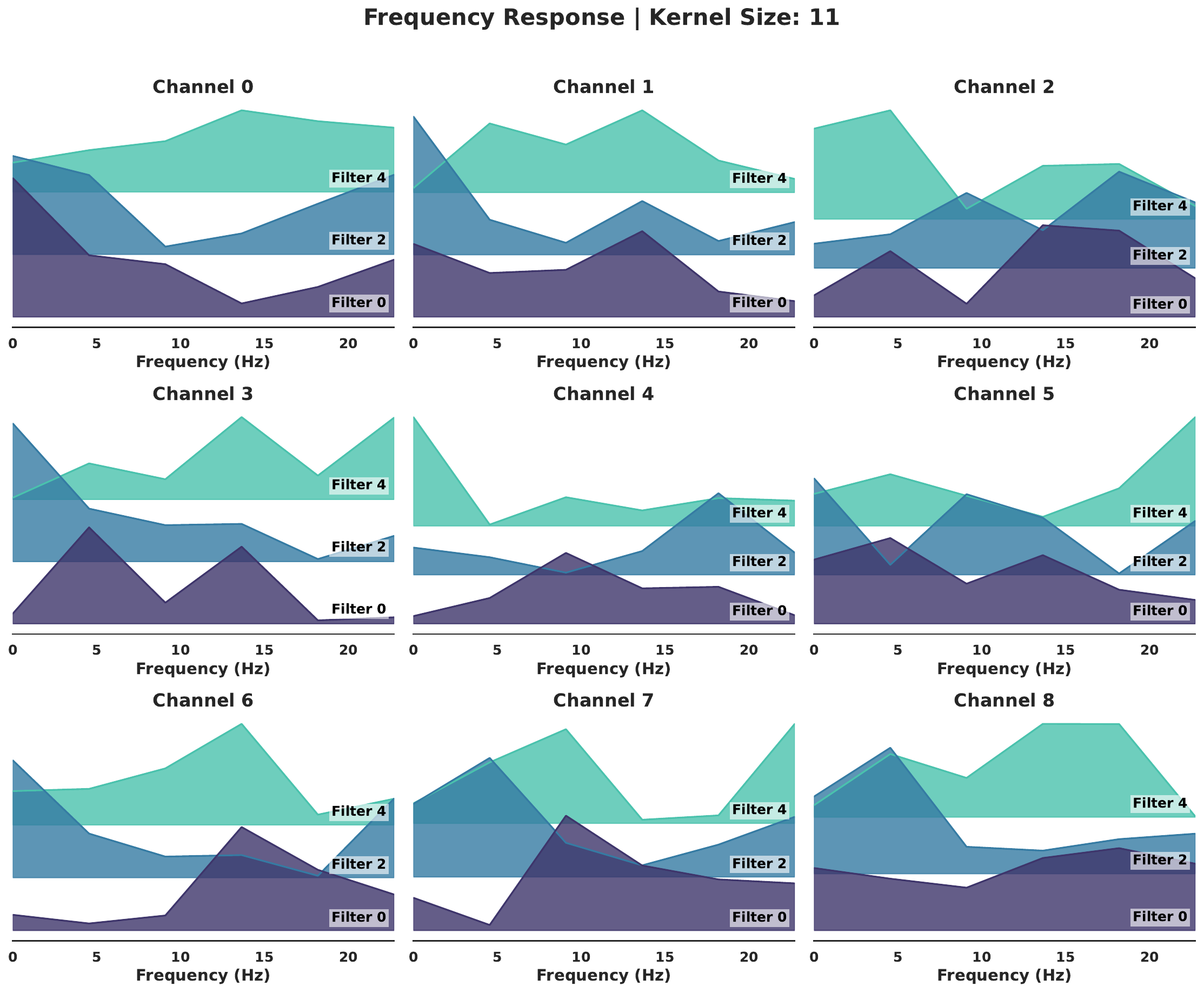}
        \caption{Kernel Size 11}
        \label{subfig:k11}
    \end{subfigure}\hfill
    \begin{subfigure}[b]{0.49\linewidth}
        \centering
        \includegraphics[width=\linewidth]{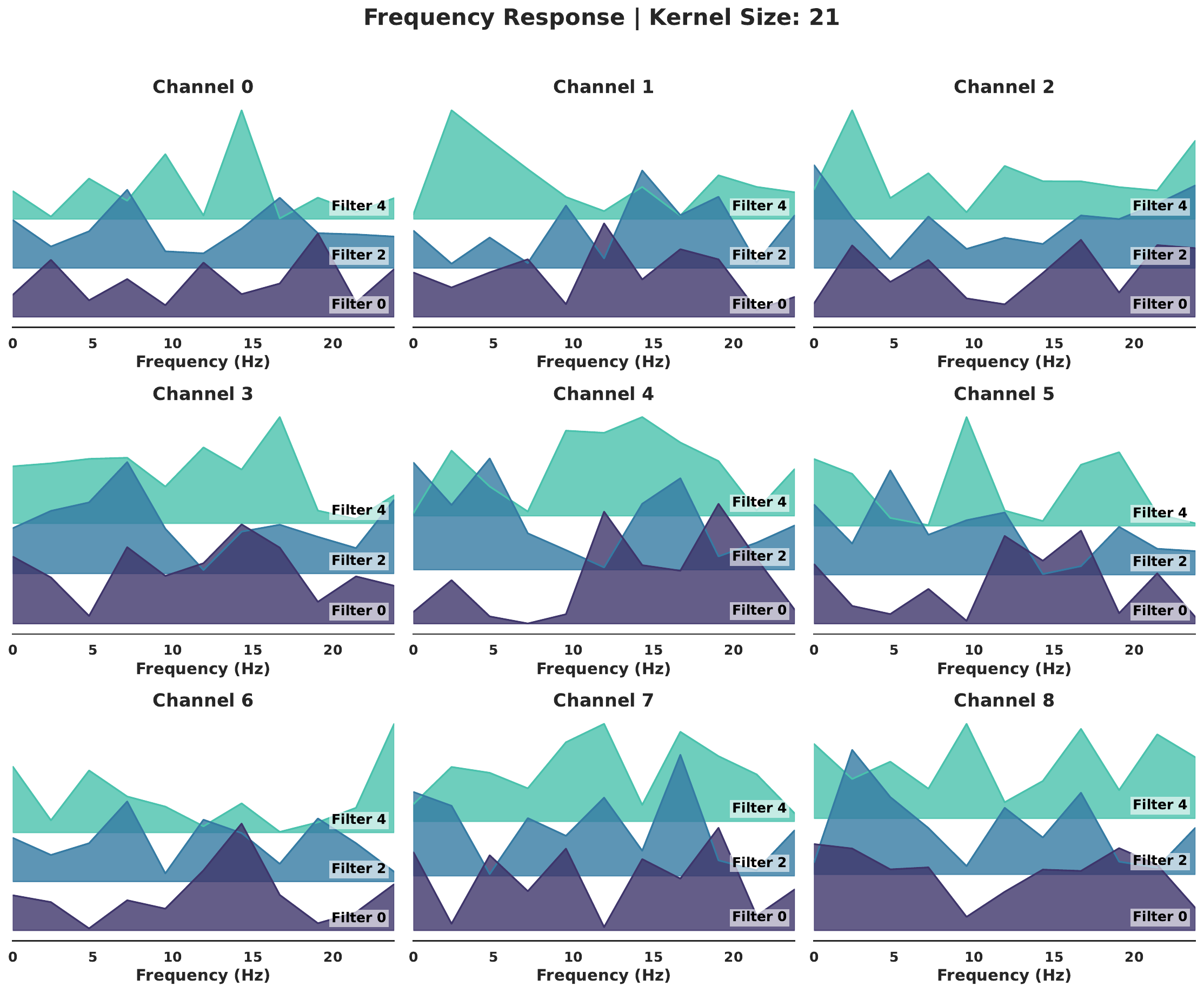}
        \caption{Kernel Size 21}
        \label{subfig:k21}
    \end{subfigure}
    
    \vspace{1em} 
    
    \begin{subfigure}[b]{0.49\linewidth}
        \centering
        \includegraphics[width=\linewidth]{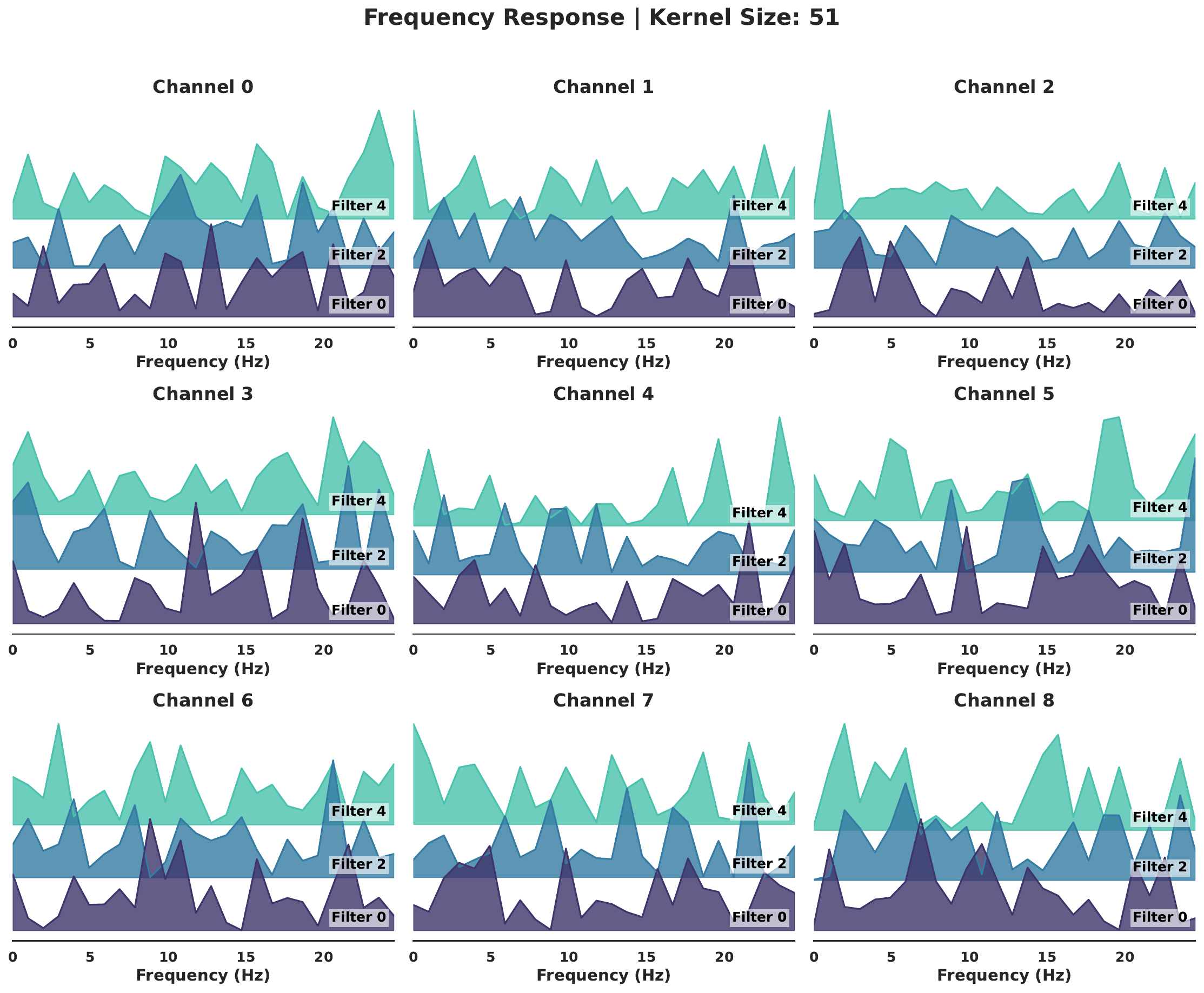}
        \caption{Kernel Size 51}
        \label{subfig:k51}
    \end{subfigure}\hfill
    \begin{subfigure}[b]{0.49\linewidth}
        \centering
        \includegraphics[width=\linewidth]{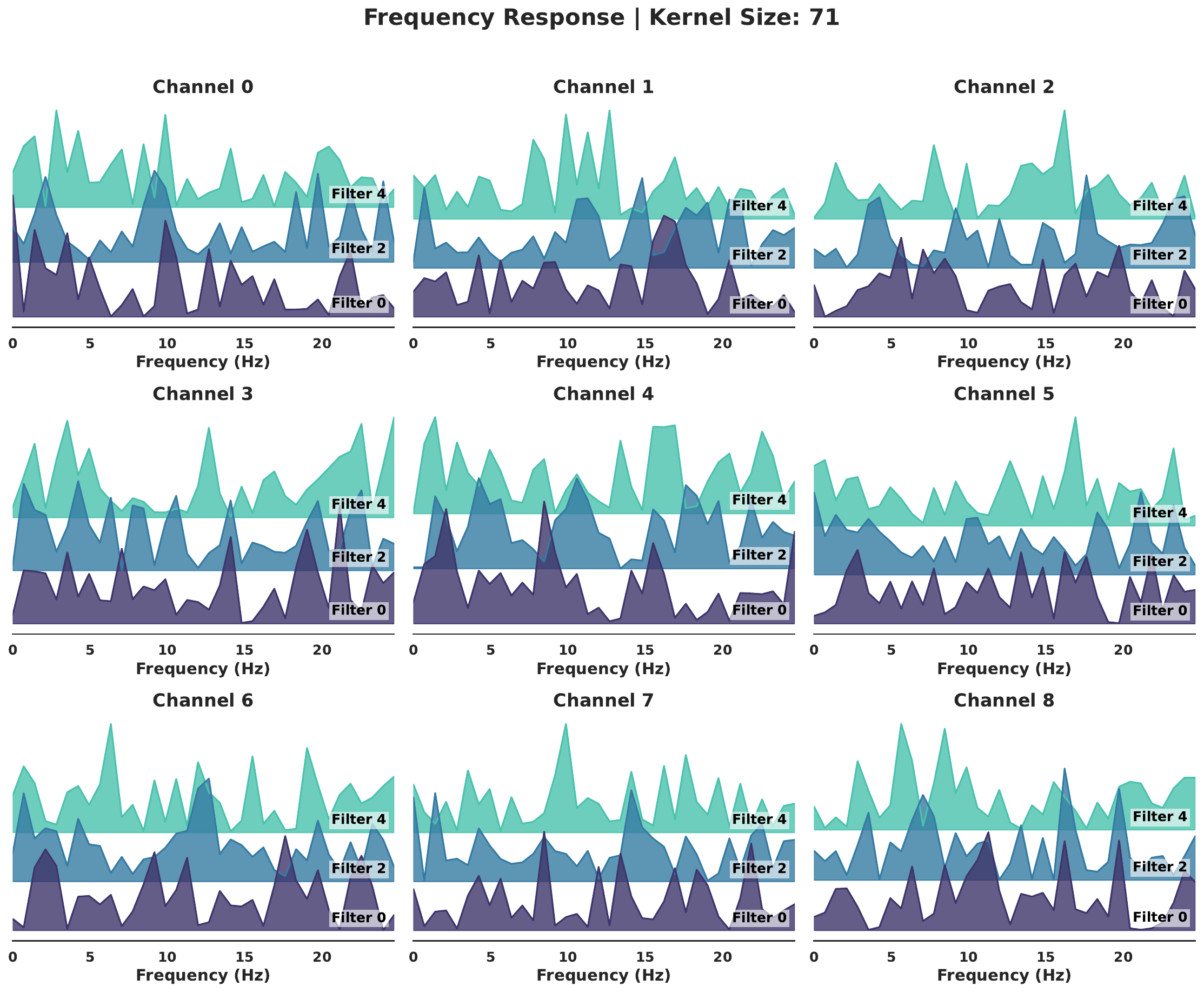}
        \caption{Kernel Size 71}
        \label{subfig:k71}
    \end{subfigure}
    
    \caption{Frequency responses of the multi-resolution filters across different channels on the UCI HAR dataset, demonstrating the shift from broad-band to highly specific frequency targeting as kernel size increases.}
    \label{fig:freq_responses}
\end{figure}

Figure \ref{fig:freq_responses}) presents the frequency responses for the CNN filters across different channels using the UCI HAR dataset. The following observation can be made:
\begin{itemize}
    \item \textbf{Channel-Specific Adaptation:} Instead of converging to uniform, generalised frequency bands across the network, the training process pushes the filters in each channel to distinct frequency profiles. This empirical divergence indicates that the model is actively adapting its receptive fields to the unique spectral characteristics of each independent input channel.
    
    \item \textbf{Role of Multi-Resolution:} By visualising different kernel sizes across the multi-resolution block, we observe that the network uses varying kernel lengths to focus on different frequency granularities. For instance, shorter kernels (e.g., size 11, Figure \ref{subfig:k11}) capture smooth, macroscopic frequency trends. As the kernel size increases (e.g., sizes 21, 51, and 71, Figures \ref{subfig:k21}, \ref{subfig:k51}, and \ref{subfig:k71}), the frequency responses become progressively more selective. These larger kernels exhibit sharp, localised peaks that isolate fine-grained spectral details. 
\end{itemize}

This qualitative analysis offers evidence that the network leverages the multi-resolution setup to combine broad-band features with highly specific frequency targeting. We hypothesise that the capacity to learn these distinct, channel-specific frequency representations contributes to the quantitative performance improvements observed in the main text's classification benchmarks, though isolating the exact causal impact remains an area for future work.

\enlargethispage*{-12pt}

\renewcommand{\journal}[1]{#1}
\providecommand{\conference}[1]{#1}
\providecommand{\book}[1]{#1}
\providecommand{\arxivtype}[1]{#1}

\providecommand{\au}[1]{#1}
\providecommand{\fnm}[1]{#1}
\providecommand{\snm}[1]{#1}
\providecommand{\atl}[1]{#1}
\providecommand{\cfname}[1]{#1}
\providecommand{\yr}[1]{#1}
\providecommand{\highlight}[1]{#1}
\providecommand{\jtl}[1]{#1}
\providecommand{\misc}[1]{#1}
\providecommand{\vol}[1]{#1}
\providecommand{\iss}[1]{#1}
\providecommand{\pg}[1]{#1}
\providecommand{\pub}[1]{#1}
\providecommand{\edn}[1]{#1}
\providecommand{\btl}[1]{#1}
\providecommand{\etal}{et~al.}
\providecommand{\arxivid}[2]{#2} 
\def\reflinktrue{}               

\end{document}